\documentclass[11pt]{article}
\PassOptionsToPackage{switch,mathlines}{lineno}
\usepackage[final]{acl}
\usepackage{times}
\usepackage{latexsym}
\usepackage[T1]{fontenc}
\usepackage[utf8]{inputenc}
\usepackage{microtype}
\usepackage{graphicx}
\usepackage{amsmath}
\usepackage{amssymb}
\usepackage{multirow}
\usepackage{booktabs}
\usepackage{xcolor}

\title{The Attribution Blind Spot: Detecting When Language Models \\
Rely on Memory Rather Than Retrieved Context}

\author{
  \textbf{Zhe Yu}$^{2*}$ \quad 
  \textbf{Wenpeng Xing}$^{1,2*}$ \quad 
  \textbf{Yunzhao Wei}$^{2}$ \quad 
  \textbf{Bo Yang}$^{3}$ \quad \\
  \textbf{Chen Ye}$^{4}$ \quad 
  \textbf{Gaolei Li}$^{5}$ \quad 
  \textbf{Meng Han}$^{1,2,6}$ \\
  \rule{0pt}{2.5ex} 
  $^{1}$Zhejiang University \quad $^{2}$Binjiang Institute of Zhejiang University \\ $^{3}$National Fintech Evaluation Center \quad $^{4}$Hangzhou Dianzi University \quad $^{5}$Shanghai Jiao Tong University \\ $^{6}$GenTel.io \\
  \rule{0pt}{1.8ex} 
  \small $^{*}$Equal contribution \\
}

\begin{document}
\maketitle

\begin{abstract}
Retrieval-augmented generation promises to ground language model outputs in external evidence, yet the field has no reliable way to verify whether retrieved context actually governs generation---a prerequisite for any high-stakes deployment. The standard assumption, that context-consistent output implies context-governed output, breaks when the retrieved document overlaps with the model's pretraining data: the model can produce faithful-looking text entirely from parametric memory, and both pathways yield indistinguishable output. We name this failure the \textbf{attribution blind spot} and introduce \textbf{Computational Reality Monitoring} (CRM) to address it. CRM operationalizes a principle adapted from cognitive science's reality monitoring framework: comparing internal representations with and without context reveals membership-conditioned representational divergence that output-level monitors systematically miss. CRM does not certify which source an individual generation used; it detects whether pretraining exposure leaves a measurable internal trajectory signature, establishing a necessary substrate for source attribution. Across nine model variants spanning three families, this divergence concentrates in architecture-specific layer patterns, receives converging support from block-level noise intervention, and generalizes across tasks and datasets while collapsing on domain-confounded benchmarks. The attribution blind spot is measurable and partially addressable: internal representations carry a diagnostic signal invisible at the output level, establishing a foundation for systems whose internal awareness of evidence provenance governs their external behavior.
\end{abstract}

\section{Introduction}

Retrieval-augmented generation~\cite{lewis2020rag,guu2020realm,borgeaud2022retro} has become a standard paradigm for grounding language model outputs in external knowledge. The operating assumption is straightforward: if a model receives a relevant document as context, it will use that document to inform its generation. This assumption underpins deployed systems in search, customer support, and medical QA, where faithful grounding is treated as a safety property.

This assumption is systematically unverifiable from outputs alone. When the retrieved document overlaps with the model's training data---common given that retrieval corpora often cover pretraining sources---the model may default to parametric memory rather than external context. The generated text appears context-grounded while the true computational pathway is parametric. We term this the \textbf{attribution blind spot}: output-level monitoring cannot distinguish read-from-context from recalled-from-parameters when both produce equally plausible text.

Prior work approaches this blind spot from the wrong level of analysis. Membership inference attacks~\cite{shokri2017mia,carlini2021extracting,shi2024min,duan2024membership} ask whether a document was seen during training---a static question, not a dynamic one about whether that document drives a specific generation. RAG faithfulness metrics~\cite{liu2024lost,niu2024ragtruth,liu2023evaluating} detect context-memory \emph{conflict} where outputs visibly contradict the provided context; our setting is harder because both sources produce identical surface text. Citation benchmarks~\cite{bohnet2022attr,gao2023ragas} evaluate whether models \emph{claim} to use context. The common blind spot: output-level signals cannot distinguish a model that read from context from one that recalled from memory when both produce identical text.

We address this gap with \textbf{Computational Reality Monitoring (CRM)}, adapted from cognitive science's reality monitoring~\cite{johnson1993reality}. Human reality monitoring distinguishes perceived from internally-generated memories by comparing sensory detail, contextual information, and cognitive operations~\cite{johnson1993reality}. CRM operationalizes this logic for language model generation: it compares the model's internal representations with the retrieved context versus without, treating the representational divergence as a diagnostic signal. The core insight: membership-conditioned differences live in the \emph{gap} between context-conditioned and unconditioned computation, not in either pathway alone.

CRM detects membership-conditioned representational divergence---whether internal states differ when the provided context is a document the model was exposed to during pretraining (member) versus one it was not (non-member). We emphasize that CRM does not certify source use for individual generations; membership is a necessary but not sufficient condition for source attribution, creating the \emph{possibility} of parametric generation, not a guarantee of it (see Section~\ref{sec:construct-validity} and Appendix~\ref{sec:caveat}). CRM establishes a measurable internal signal that future source-attribution systems could build upon---a diagnostic substrate, not a final verifier. Our contributions are:
\begin{enumerate}
    \item \textbf{The attribution blind spot:} We formalize the failure mode where parametric memory and retrieved context agree on the surface, making output-level source attribution impossible.
    \item \textbf{Architecture-dependent layer localization with causal evidence:} Across nine model variants, membership-conditioned signals localize non-monotonically in three architecture-dependent patterns (bimodal, mid-layer, scattered-late). Block-level noise injection provides causal evidence that CRM-identified blocks contribute to preserving membership-conditioned information, validating a distributed-encoding hypothesis for architectures where single-layer perturbation had no effect.
    \item \textbf{Direction design and evidentiary bounds:} A supervised mean-difference direction universally improves over unsupervised PC1 ($\Delta$AUC $+$0.024--0.144). CRM generalizes across tasks (summarization, QA) and datasets (BookMIA AUC 0.84--0.97), withstands same-topic control, and collapses on domain-confounded benchmarks (MIMIR), establishing boundary conditions. CRM-LTS is competitive with gradient-based, attention-based, and logit-lens baselines while uniquely supporting layer-localized causal interpretation.
    \item \textbf{Deployment prototype:} A FastAPI audit server with real-time trajectory dashboard demonstrates CRM's compact scalar-per-layer signature enables low-latency deployment auditing.
\end{enumerate}

\section{Computational Reality Monitoring}

\paragraph{Problem formulation.} Let $\mathcal{M}$ be a model with parameters $\theta$ pretrained on $\mathcal{D}_{\text{train}}$. For query $q$ with retrieved context $c$, $\mathcal{M}$ produces $y_0 = \mathcal{M}(q)$ (no-context) and $y_c = \mathcal{M}(c, q)$ (with-context). CRM detects whether $c \in \mathcal{D}_{\text{train}}$ (member-conditioned) versus $c \notin \mathcal{D}_{\text{train}}$ (non-member-conditioned) by comparing internal representations across these two conditions. Membership is an experimental proxy that creates the \emph{possibility} of parametric generation (Appendix~\ref{sec:caveat}).

\paragraph{Three-level framework.} CRM examines internal representations at three levels. Level~1 (black-box) measures BGE-M3~\cite{bge-m3} embedding distance between with/no-context generations~\cite{reimers2019sbert}. Level~2 (grey-box) computes per-step KL divergence from the LM head, aggregated into five statistics. Level~3 (white-box), CRM's core, probes hidden states.

\paragraph{Latent Trajectory Shift (LTS).} For target layers $\mathcal{L}$ selected by PCA variance ratio ($>$0.01), we extract hidden states $h_\ell^0$ and $h_\ell^c$ at the last token position. We reserve $n_{\text{cal}}=100$ samples for computing PC1 directions $v_\ell$ via SVD on displacement vectors $d_\ell = h_\ell^c - h_\ell^0$; all $N=250$ samples are then used for feature extraction and evaluation under 5-fold stratified CV. The signed scalar projection
\begin{equation}
    \text{LTS}_\ell = \langle h_\ell^c - h_\ell^0,\; v_\ell \rangle
\end{equation}
captures directional displacement along $v_\ell$. For the supervised direction $v_{\text{sup}}$ (Section~\ref{sec:supervised-direction}), the same calibration subset is used to compute the mean-difference direction. Using PC1 per layer yields 9--22 compact trajectory features. The unified feature vector $\Phi = [\Phi_{\text{L1}}, \Phi_{\text{L2}}, \Phi_{\text{L3}}]$ is evaluated with logistic regression~\cite{pedregosa2011sklearn} and XGBoost~\cite{chen2016xgboost} under 5-fold stratified CV. Full equations and L1/L2 definitions in Appendix~\ref{sec:crm-equations}.

\begin{figure*}[t]
  \centering
  \includegraphics[width=\textwidth]{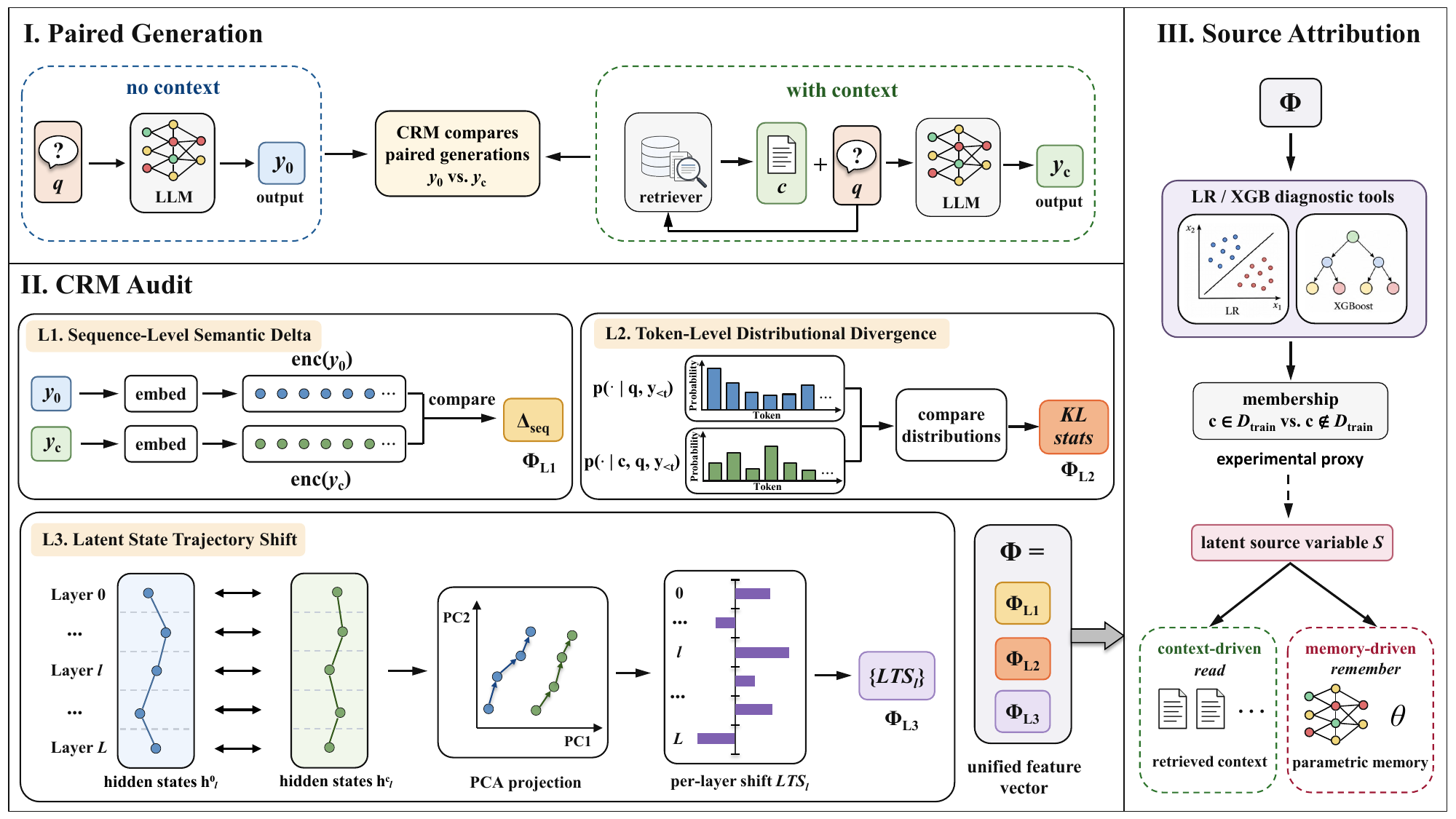}
  \caption{Computational Reality Monitoring framework. CRM compares paired no-context and with-context generations, extracts sequence-, token-, and latent-level divergence features, and uses the resulting feature vector to detect membership-conditioned representational divergence---a diagnostic signal for whether pretraining exposure reshapes the model's internal computation, not a per-generation source verifier.}
  \label{fig:crm-framework}
\end{figure*}

\section{Experimental Setup}
\label{sec:setup}

We adopt a controlled diagnostic design: continuation probing isolates context-memory interaction and eliminates confounds such as query formulation and instruction following.

\textbf{Models and data.} Nine Transformer~\cite{vaswani2017attention} variants: Llama-3.1-8B/Instruct~\cite{meta2024llama3}, Mistral-7B-v0.3/Instruct~\cite{jiang2023mistral}, and Qwen2.5 (7B/7B-Inst/14B/14B-Inst/32B-Inst)~\cite{yang2024qwen2}. WikiMIA~\cite{shi2024min}: 250 balanced samples/model (128-token passages; members from Wikipedia dumps before 2017-03-20, non-members from after 2018-02-01). Cross-dataset: BookMIA~\cite{shi2024min} (Books3 domain split; Appendix~\ref{sec:bookmia}). Negative control: MIMIR Pile-Wikipedia split~\cite{duan2024membership} (Appendix~\ref{sec:mimir}). Target layers $\mathcal{L}$ are selected by PCA variance ratio $>$0.01 on calibration-set displacement vectors, filtering out near-zero-variance layers; this yields 9--22 layers per model (Table~\ref{tab:main}, L3 dim column).

\textbf{Baselines.} Three tiers: (1)~Black-box likelihood (PPL, Zlib-PPL~\cite{carlini2021extracting}, Min-K\% Prob~\cite{shi2024min}); (2)~Access-matched (single-layer LTS, mean LTS, L1+L2 only); (3)~Raw hidden-state probes. Full details: Appendix~\ref{sec:baseline-details}.

\textbf{Evaluation.} 5-fold stratified CV (seed 42), ROC-AUC with 95\% bootstrap CIs. Controls: label permutation, prompt randomization (4 templates), same-topic (BGE-M3 similarity-matched non-members, $n{\approx}140$). Methodology: Appendix~\ref{sec:appendix-controls}.

\section{Results}

The attribution blind spot predicts a specific empirical signature: if output-level monitors cannot distinguish generation conditions when both pathways produce plausible text, then the discriminative signal must reside elsewhere---in the model's internal computation. CRM directly tests this prediction by comparing representations with and without context. We present five lines of evidence: (i)~CRM consistently separates member-conditioned from non-member-conditioned generation while surface baselines remain near chance; (ii)~the signal is latent-dominated, with L1+L2 contributing negligibly; (iii)~the signal survives same-topic control, label permutation, and prompt randomization; (iv)~membership-conditioned divergence localizes in architecture-dependent layer patterns; and (v)~directional displacement (CRM-LTS) and isotropic magnitude (L2) carry qualitatively distinct information.

\subsection{Main Results}

\begin{table*}[t]
\centering
\small
\begin{tabular}{lccccc}
\toprule
\textbf{Model} & \textbf{Best Likelihood BL} & \textbf{CRM-LR} & \textbf{CRM-XGB} & \textbf{Gain} & \textbf{L3 dim} \\
\midrule
Llama-3.1-8B        & 0.565 & 0.778 [0.741, 0.838] & 0.707 [0.655, 0.759] & +0.213 & 12 \\
Llama-3.1-8B-Inst   & 0.575 & 0.708 [0.665, 0.760] & 0.627 [0.553, 0.687] & +0.133 & 12 \\
Mistral-7B-v0.3     & 0.575 & 0.869 [0.843, 0.890] & 0.815 [0.777, 0.856] & +0.294 & 12 \\
Mistral-7B-Inst     & 0.596 & 0.799 [0.744, 0.845] & 0.731 [0.680, 0.769] & +0.203 & 12 \\
Qwen2.5-7B          & 0.583 & 0.784 [0.765, 0.801] & 0.765 [0.739, 0.791] & +0.201 & 10 \\
Qwen2.5-7B-Inst     & 0.579 & 0.869 [0.837, 0.901] & 0.815 [0.777, 0.860] & +0.290 & 10 \\
Qwen2.5-14B         & 0.562 & 0.840 [0.793, 0.874] & 0.777 [0.756, 0.797] & +0.278 & 17 \\
Qwen2.5-14B-Inst    & 0.567 & \textbf{0.951} [0.941, 0.961] & 0.864 [0.842, 0.885] & \textbf{+0.384} & 17 \\
Qwen2.5-32B-Inst    & 0.550 & 0.923 [0.881, 0.957] & 0.880 [0.815, 0.923] & +0.373 & 23 \\
\midrule
\textit{Mean}       & 0.572 & 0.836 & 0.778 & +0.264 & -- \\
\bottomrule
\end{tabular}
\caption{\textbf{Likelihood-based baselines fail to separate source conditions (AUC 0.55--0.60), while CRM consistently distinguishes member-conditioned from non-member-conditioned generation across all nine models (AUC 0.71--0.95).} Best Likelihood BL = highest AUC among three Tier-1 token-likelihood baselines (PPL, Zlib-PPL, Min-K\% Prob), all operating on document text alone \emph{without the generation contrast} that CRM exploits. 95\% bootstrap CIs in brackets. Gain = CRM-LR $-$ Best Likelihood BL. L3 dim = number of target-layer LTS features.}
\label{tab:main}
\end{table*}

Table~\ref{tab:main} shows the main results. \textbf{Baselines remain near chance} (AUC 0.55--0.60), confirming token-level memorization signals are weak in modern LLMs. \textbf{CRM consistently separates conditions} (AUC 0.71--0.95, gain +0.13 to +0.38), with a family-level gradient: Qwen (mean 0.873) $>$ Mistral (0.834) $>$ Llama (0.743). Logistic regression matches or exceeds XGBoost, aligning with findings on linear representation of high-level concepts~\cite{park2024linear,zou2023linear,templeton2024scaling,bricken2023monosemanticity}.

\subsection{Latent Signal and Robustness}

Table~\ref{tab:l3} reports a defining empirical finding: removing all surface features (L1+L2) changes LR AUC by less than 0.01 across nine models (mean $|\Delta|=0.006$). Membership-conditioned divergence is almost entirely latent---output-accessible signals alone are insufficient for detection.

\begin{table}[t]
\centering
\small
\begin{tabular}{lccc}
\toprule
\textbf{Model} & \textbf{Full CRM} & \textbf{L3-Only} & $\Delta$LR \\
\midrule
Llama-3.1-8B        & 0.779 & 0.781 & $-$0.002 \\
Llama-3.1-8B-Inst   & 0.709 & 0.701 & +0.009 \\
Mistral-7B-v0.3     & 0.869 & 0.860 & +0.009 \\
Mistral-7B-Inst     & 0.800 & 0.777 & +0.023 \\
Qwen2.5-7B          & 0.786 & 0.799 & $-$0.013 \\
Qwen2.5-7B-Inst     & 0.870 & 0.872 & $-$0.003 \\
Qwen2.5-14B         & 0.841 & 0.835 & +0.006 \\
Qwen2.5-14B-Inst    & 0.950 & 0.947 & +0.004 \\
Qwen2.5-32B-Inst    & 0.922 & 0.927 & $-$0.005 \\
\midrule
\textit{Mean}       & 0.836 & 0.833 & +0.003 \\
\bottomrule
\end{tabular}
\caption{\textbf{Removing surface features (L1+L2) changes AUC by $<$0.01---the signal is latent.} Full results: Appendix~\ref{sec:access-app}.}
\label{tab:l3}
\end{table}

\begin{figure}[t]
\centering
\includegraphics[width=\columnwidth]{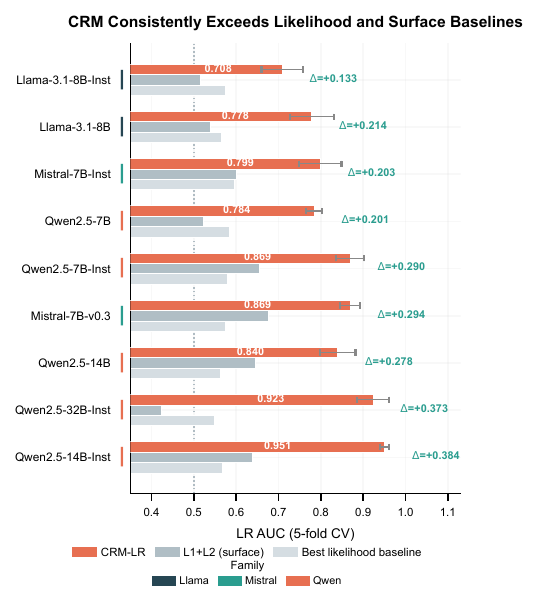}
\caption{\textbf{CRM consistently exceeds likelihood and surface baselines.} CRM-LR AUC (coral), L1+L2 surface features (gray), best likelihood baseline (light gray). Error bars: 95\% CI.}
\label{fig:main-results}
\end{figure}

\begin{table*}[t]
\centering
\small
\begin{tabular}{lcccc}
\toprule
\textbf{Model} & \textbf{Prompt Std} & \textbf{Permuted AUC} & \textbf{ST CRM-LR} & $\Delta$ vs Rand \\
\midrule
Qwen2.5-14B-Inst & 0.019 & 0.50 $\pm$ 0.05 & 0.921 $\pm$ 0.033 & $-$0.004 \\
Mistral-7B-v0.3  & 0.009 & 0.50 $\pm$ 0.05 & 0.842 $\pm$ 0.060 & +0.020 \\
Llama-3.1-8B     & 0.011 & 0.50 $\pm$ 0.05 & 0.726 $\pm$ 0.072 & $-$0.058 \\
\bottomrule
\end{tabular}
\caption{\textbf{CRM's diagnostic signal withstands three robustness challenges.} Prompt variation: AUC std $<$ 0.02 across four templates. Label permutation: AUC returns to 0.50 $\pm$ 0.05. Same-topic control (CRM-LTS, $n{\approx}140$): AUC largely preserved ($\Delta$ within $\pm$0.06) despite 1.6$\times$ tighter topic matching. ST CRM-LR = same-topic CRM-LTS logistic regression AUC. Full methodology and per-template prompt breakdown in Appendix~\ref{sec:appendix-controls}.}
\label{tab:robustness}
\end{table*}

\begin{figure*}[t]
\centering
\includegraphics[width=\textwidth]{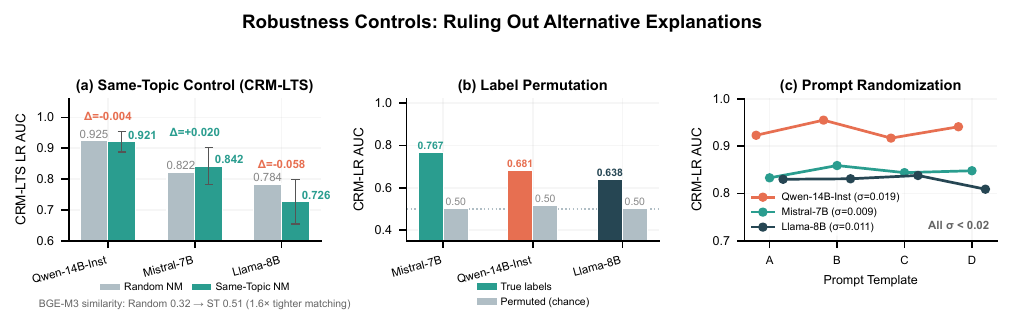}
\caption{\textbf{Robustness controls rule out topic familiarity, classifier artifacts, and prompt artifacts.} (a) Same-topic control (CRM-LTS): AUC largely preserved ($\Delta$ within $\pm$0.06). (b) Label permutation: AUC returns to chance. (c) Prompt randomization: $\sigma < 0.02$. Dashed line: AUC = 0.50.}
\label{fig:robustness}
\end{figure*}

Three robustness checks rule out alternative explanations (Table~\ref{tab:robustness}; full methodology: Appendix~\ref{sec:appendix-controls}). \textbf{Same-topic control:} semantically similar non-members (BGE-M3 similarity 0.51 vs.\ 0.32) yield CRM-LTS AUC within $\pm$0.06 across three models (Qwen: $-$0.004, Mistral: +0.020, Llama: $-$0.058), ruling out topic familiarity. \textbf{Label permutation} returns AUC to 0.50 $\pm$ 0.05. \textbf{Prompt randomization} across four templates yields AUC std $<$0.02. A cross-task pilot replacing continuation with summarization preserves the CRM signal for Qwen ($\Delta=+0.007$) with partial transfer for Mistral ($\Delta=-0.125$, AUC 0.744; Appendix~\ref{sec:summ-pilot}).

\subsection{Layer-wise Analysis}
\label{sec:layers}

\begin{table*}[t]
\centering
\small
\begin{tabular}{lccccl}
\toprule
\textbf{Model} & \textbf{Layers} & \textbf{Best L} & \textbf{LR AUC} & \textbf{XGB AUC} & \textbf{Pattern} \\
\midrule
Qwen2.5-14B-Inst  & 48 & L6 / L21 & 0.843 & 0.732 / 0.716 & Bimodal (early + mid-deep) \\
Qwen2.5-7B        & 28 & L10      & \textbf{0.902} & \textbf{0.894} & Early-mid peak \\
Mistral-7B-v0.3   & 32 & L18      & 0.892 & 0.795 & Mid-layer peak \\
Llama-3.1-8B      & 32 & L28      & 0.753 & 0.653 & Scattered late \\
\bottomrule
\end{tabular}
\caption{\textbf{Membership-conditioned divergence is non-monotonic and architecture-dependent.} Single-layer LTS AUC peaks at different stages depending on model family. The Qwen-14B-Inst bimodal pattern (peaks at L6 and L21, trough at L27 below chance) shows divergence appearing, disappearing, and reappearing---inconsistent with uniform depth-based hypotheses. 50-sample diagnostic subsets; values not directly comparable to 250-sample multi-layer results.}
\label{tab:layers}
\end{table*}

Table~\ref{tab:layers} summarizes the layer sweep (full per-layer breakdown: Appendix~\ref{sec:per-layer-app}). Membership-conditioned divergence is \textbf{non-monotonically related to depth}, exhibiting three architecture-dependent patterns: \textbf{bimodal} (Qwen2.5-14B-Inst: peaks at L6/L21, trough at L27 below chance), \textbf{mid-layer concentration} (Mistral-7B: L18/0.892; Qwen2.5-7B: L10/0.902), and \textbf{scattered late} (Llama-3.1-8B: L28/0.753, distributed). The below-chance trough at Qwen L27 suggests active suppression. Leave-one-layer-out ablation confirms no single layer is uniquely informative ($|\Delta|$AUC $<$ 0.007; Appendix~\ref{sec:loo}), indicating redundant encoding.

\begin{figure*}[t]
\centering
\includegraphics[width=\textwidth]{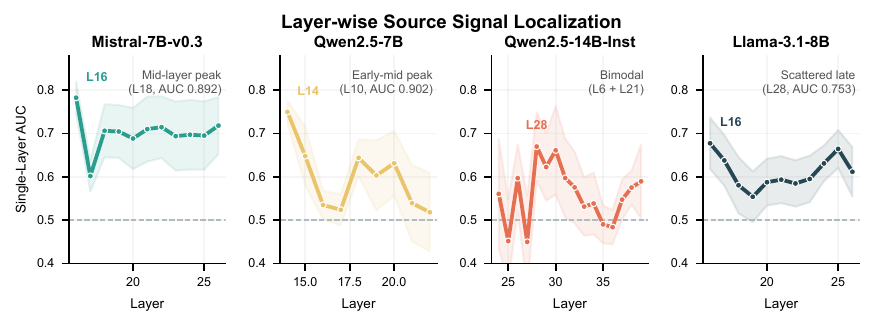}
\caption{\textbf{Layer-wise CRM-LTS single-probe AUC reveals architecture-dependent patterns.} Qwen bimodal (L6/L21), Mistral mid-layer (L18), Qwen-7B early (L10), Llama scattered-late (L28). Bands: $\pm$1 std. Dashed: AUC = 0.50.}
\label{fig:layer-sweep}
\end{figure*}

\subsection{Causal Evidence: Block-Level Noise Injection}
\label{sec:causal}

Single-layer noise injection (Appendix~\ref{sec:single-layer-noise}) revealed a puzzle: Qwen2.5-14B-Inst L6 showed catastrophic degradation ($\Delta$AUC $-$0.300), but Llama-3.1-8B L28 was near-immune ($-$0.001). The \textbf{distributed-encoding hypothesis} resolves this: membership information is spread across a block such that perturbing any single layer leaves sufficient signal elsewhere for recovery. We test this by simultaneously injecting noise ($\mathbf{h}' = \mathbf{h} + \varepsilon \cdot \sigma(\mathbf{h}) \cdot \mathcal{N}(0, 1)$) across architecture-specific blocks informed by per-layer AUC patterns (Table~\ref{tab:layers}): Llama \texttt{scattered\_late} L25--L31 (all layers AUC$>$0.5), Mistral \texttt{mid\_cluster} L14--L22 ($\pm$4 around L18), Qwen \texttt{early\_peak} L4--L8 and \texttt{late\_peak} L19--L23 (the two bimodal peaks). \texttt{early\_control} L0--L7/8 serves as a universal baseline.

\begin{table}[t]
\centering
\footnotesize
\setlength{\tabcolsep}{3pt}
\resizebox{\columnwidth}{!}{%
\begin{tabular}{llccc}
\toprule
\textbf{Model} & \textbf{Block (layers)} & \textbf{$\varepsilon$} & \textbf{AUC} & $\Delta$AUC \\
\midrule
\multirow{4}{*}{Llama-8B} & \multirow{2}{*}{\texttt{scattered\_late} (25--31)} & 0.1 & 0.956 & $-$0.001 \\
 & & 0.5 & 0.936 & \textbf{$-$0.021} \\
 & \multirow{2}{*}{\texttt{early\_control} (0--7)} & 0.1 & 0.570 & $-$0.387 \\
 & & 0.5 & 0.552 & $-$0.405 \\
\midrule
\multirow{4}{*}{Mistral-7B} & \multirow{2}{*}{\texttt{mid\_cluster} (14--22)} & 0.1 & 0.800 & $-$0.025 \\
 & & 0.5 & 0.780 & \textbf{$-$0.045} \\
 & \multirow{2}{*}{\texttt{early\_control} (0--7)} & 0.1 & 0.540 & $-$0.285 \\
 & & 0.5 & 0.565 & $-$0.260 \\
\midrule
\multirow{4}{*}{Qwen-14B-Inst} & \multirow{2}{*}{\texttt{early\_peak} (4--8)} & 0.1 & 0.540 & \textbf{$-$0.408} \\
 & & 0.5 & 0.557 & $-$0.391 \\
 & \multirow{2}{*}{\texttt{late\_peak} (19--23)} & 0.1 & 0.915 & $-$0.032 \\
 & & 0.5 & 0.893 & $-$0.054 \\
\midrule
\multicolumn{5}{c}{\textbf{Clean baselines:} Llama 0.957, Mistral 0.825, Qwen 0.947} \\
\bottomrule
\end{tabular}%
}
\caption{\textbf{Block-level noise injection provides causal evidence that CRM-identified blocks contribute to preserving membership-conditioned information.} Target blocks show selective degradation; early control blocks produce catastrophic collapse.}
\label{tab:causal-block}
\end{table}

Table~\ref{tab:causal-block} reports four findings. \textbf{(1) Distributed-encoding supported:} Llama's \texttt{scattered\_late} block---selected because single-layer L28 showed near-zero effect---shows measurable degradation under simultaneous perturbation ($\Delta$AUC $-$0.021 at $\varepsilon$=0.5, $\sim$20$\times$ the single-layer effect). \textbf{(2) Architecture-dependent selectivity:} Qwen's \texttt{early\_peak} catastrophically collapses ($-$0.408) while \texttt{late\_peak} degrades mildly ($-$0.032 to $-$0.054), confirming the bimodal peaks serve distinct causal roles. Mistral's \texttt{mid\_cluster} shows the strongest relative target-block effect ($-$0.045). \textbf{(3) Early layers are causally foundational:} \texttt{early\_control} produces catastrophic collapse across all three models ($-$0.26 to $-$0.41), providing a positive control. \textbf{(4) Why patching fails:} Activation patching---replacing $h_\ell^c$ at the last token with $h_\ell^0$---yields negligible $\Delta$AUC ($<$0.01). Residual connections allow downstream recovery after single-position patching; noise injection perturbs all token positions simultaneously, preventing recovery.

These results indicate that CRM-identified layers are not merely diagnostic but contribute to carrying membership-conditioned information. The early-control collapse demonstrates that early-layer computation is foundational to all downstream representation, making it a weaker test for selective encoding. The architecture-dependent target-block effect sizes mirror the layer-localization patterns (Table~\ref{tab:layers}), and the distributed-encoding hypothesis receives direct support.

\subsection{Supervised vs.\ Unsupervised Direction}
\label{sec:supervised-direction}

CRM-LTS projects displacement vectors onto PC1---an unsupervised direction maximizing explained variance. PCA optimizes reconstruction, not discriminability. We test a supervised alternative: for each layer $\ell$, compute the mean-difference direction
\begin{equation}
    v_{\text{sup}}^{(\ell)} = \frac{\mathbb{E}[d_\ell \mid \text{member}] - \mathbb{E}[d_\ell \mid \text{non-member}]}{\|\mathbb{E}[d_\ell \mid \text{member}] - \mathbb{E}[d_\ell \mid \text{non-member}]\|_2}
\end{equation}
and define $\text{LTS}_{\text{sup}}^{(\ell)} = \langle h_\ell^c - h_\ell^0,\; v_{\text{sup}}^{(\ell)} \rangle$. Both directions use the same 100-sample calibration set and 5-fold CV.

\begin{table}[t]
\centering
\footnotesize
\setlength{\tabcolsep}{3.2pt}
\resizebox{\columnwidth}{!}{%
\begin{tabular}{lcccc}
\toprule
\textbf{Model} & \textbf{PC1 AUC} & \textbf{Sup. AUC} & $\Delta$AUC & \textbf{Layers} \\
\midrule
Llama-3.1-8B      & 0.957 $\pm$ 0.019 & 0.981 $\pm$ 0.013 & \textbf{+0.024} & 32 \\
Mistral-7B-v0.3   & 0.825 $\pm$ 0.030 & 0.969 $\pm$ 0.015 & \textbf{+0.144} & 32 \\
Qwen2.5-14B-Inst  & 0.947 $\pm$ 0.019 & 0.974 $\pm$ 0.015 & \textbf{+0.027} & 48 \\
\bottomrule
\end{tabular}%
}
\caption{\textbf{Supervised mean-difference direction universally outperforms unsupervised PC1.} Both directions use identical per-layer LTS projection; only the projection vector differs. $\Delta$AUC = Supervised $-$ PC1. Mistral-7B shows the largest gain ($+$0.144), confirming that PCA variance maximization can substantially misalign with the discriminative direction. Even for Llama and Qwen, where PC1 already achieves strong AUC, the supervised direction provides a consistent improvement. $\pm$ = 1 std across CV folds.}
\label{tab:supervised-direction}
\end{table}

Table~\ref{tab:supervised-direction} reports the comparison on three models. \textbf{The supervised direction universally improves over PC1} ($\Delta$AUC $+$0.024--0.144), with architecture-dependent gains: Mistral-7B gains $+$0.144 (17\% relative), while Llama and Qwen gain $+$0.024--0.027. This mirrors the layer-localization typology---Mistral's mid-layer-concentrated signal is most misaligned with PC1, while Llama's distributed pattern happens to align PC1 closer to the discriminative direction. We retain PC1 as the default for its unsupervised property (no membership labels needed), noting that access to calibration labels unlocks a consistently stronger direction. This result also validates the PC rank finding (Appendix~\ref{sec:pca-sweep}): PCA does not optimize for the membership/non-membership axis.

\subsection{Cross-Task and Cross-Dataset Generalization}
\label{sec:multitask}

CRM generalizes across generation tasks. Table~\ref{tab:multitask} evaluates CRM-LTS on continuation, summarization, and factoid QA across six models. \textbf{Signal strength is task-dependent:} Mistral-family models show amplified discriminability under summarization ($\Delta=+0.063$ to $+0.078$), while Qwen2.5-14B-Instruct shows near-equal performance (0.948--0.967). If CRM only detected ease of next-token prediction, summarization---which requires deeper integration---should show weaker, not stronger, discriminability. \textbf{QA preserves strong signal across all architectures} (AUC 0.83--0.97), ruling out continuation-specific artifacts.

\begin{table}[t]
\centering
\footnotesize
\setlength{\tabcolsep}{3pt}
\begin{tabular}{lcccc}
\toprule
\textbf{Model} & \textbf{Cont.} & \textbf{Summ.} & \textbf{QA} & $\Delta$Summ \\
\midrule
Mistral-7B-v0.3   & 0.825 & 0.902 & 0.830 & \textbf{+0.078} \\
Mistral-7B-Inst   & 0.788 & 0.851 & 0.867 & \textbf{+0.063} \\
Qwen2.5-14B-Inst  & 0.948 & 0.964 & 0.967 & \textbf{+0.016} \\
Qwen2.5-7B-Inst   & 0.969 & 0.956 & 0.955 & $-$0.013 \\
Llama-3.1-8B      & 0.957 & 0.867 & 0.863 & $-$0.090 \\
Llama-3.1-8B-Inst & 0.828 & 0.815 & 0.862 & $-$0.013 \\
\bottomrule
\end{tabular}
\caption{\textbf{CRM generalizes across generation tasks.} CRM-LTS LR AUC, 5-fold CV, 250 samples. $\Delta$Summ = Summarization $-$ Continuation. Mistral-family models show summarization amplification. QA preserves AUC $>$0.83 across all architectures. Full breakdown: Appendix~\ref{sec:multitask-app}.}
\label{tab:multitask}
\end{table}

CRM also generalizes across datasets. On BookMIA~\cite{shi2024min}---where membership is defined by domain (books in vs.\ out of Books3)---CRM-LTS achieves AUC 0.84--0.97 (Table~\ref{tab:bookmia}), matching or exceeding WikiMIA performance. QA format amplifies the signal further (AUC 0.96--0.98). In contrast, MIMIR's Pile-Wikipedia split~\cite{duan2024membership}---where membership is confounded with corpus origin---yields chance-level AUC (0.48--0.55; Appendix~\ref{sec:mimir}). This failure reveals a boundary condition: CRM requires member and non-member populations from comparable distributions; when the split is domain-driven, the signal is undetectable.

\begin{table*}[t]
\centering
\small
\begin{tabular}{lccccc}
\toprule
\textbf{Model} & \textbf{WikiMIA (CRM)} & \textbf{BookMIA (CRM)} & \textbf{BookMIA (L2)} & \textbf{BookMIA-QA (CRM)} \\
\midrule
Qwen2.5-14B-Inst & 0.948 & 0.844 & 0.683 & \textbf{0.980} \\
Mistral-7B-v0.3  & 0.825 & \textbf{0.967} & 0.823 & \textbf{0.959} \\
Llama-3.1-8B     & 0.957 & 0.905 & 0.745 & \textbf{0.969} \\
\bottomrule
\end{tabular}
\caption{\textbf{CRM generalizes across datasets.} CRM-LTS LR AUC, 5-fold CV. CRM-LTS outperforms L2 on BookMIA ($\Delta = +0.09$ to $+0.16$). BookMIA-QA uses QA prompts. Full results: Appendix~\ref{sec:bookmia}.}
\label{tab:bookmia}
\end{table*}

\subsection{Additional Evidence and Boundary Conditions}

We evaluate several alternative interpretations and establish evidentiary bounds (full results in appendices). \textbf{Static MIA vs.\ generation contrast:} CRM-LR exceeds standard MIA (final-layer document embeddings, 4,096--5,120 dims) on 7/9 models by $+$0.05--0.18 with 200--500$\times$ fewer features, confirming generation contrast adds signal beyond static membership (Appendix~\ref{sec:mia-app}). For Llama, MIA exceeds CRM, but raw probes confirm abundant membership-conditioned signal (AUC 0.93--0.95; Appendix~\ref{sec:hidden-probes-app})---CRM-LTS is a lossy summary for Llama's distributed geometry.

\textbf{Comparison with attribution baselines.} CRM-LTS is competitive with gradient norm (chance, AUC 0.500), attention flow (mean 0.818), and logit lens (mean 0.859) while uniquely supporting layer-localized causal interpretation (Table~\ref{tab:competing} in Appendix~\ref{sec:competing-app}). Logit lens is the strongest competitor (Qwen: 0.953), but provides no mechanism for identifying \emph{which} computational stage encodes membership.

\textbf{L2 norm vs.\ directional displacement.} Isotropic magnitude ($\|\mathbf{h}_\ell^c - \mathbf{h}_\ell^0\|_2$) achieves comparable mean AUC (0.812 vs.\ CRM 0.836) with identical dimensionality. However, L2 Multilayer exceeds CRM-LTS on 4/9 models (Llama-8B-Inst, Mistral-7B-v0.3, Qwen2.5-7B, Qwen2.5-14B; $\Delta$ $-$0.006 to $-$0.060), revealing that directional compression is not uniformly beneficial---for some architectures, magnitude alone is a more discriminative membership signal. Critically, L2 and CRM peak at different layers (L2 early, CRM mid-to-late) and diverge under same-topic control: Mistral-7B L2 degrades from 0.929 to 0.580 while CRM-LTS is preserved (Appendix~\ref{sec:l2-app}). L2 and CRM capture qualitatively distinct signals---magnitude vs.\ directional displacement---and their concatenation is complementary on 5/9 models.

\textbf{PC1 interpretation.} PC rank ablation reveals PC1 is not the optimal discriminative direction: Mistral-7B's PC5 outperforms PC1 by 0.138 AUC (Appendix~\ref{sec:pca-sweep}). PC1 vocabulary back-projection through the LM head reveals layer-dependent semantic progression: early-layer PC1 captures subword fragments, late-layer PC1 converges to epistemic-stance markers (\texttt{Unfortunately}, \texttt{regret} for Qwen; \texttt{specific}, \texttt{exact} for Mistral; Appendix~\ref{sec:pc1-app}).

\section{Discussion}

CRM is not designed for maximal AUC---raw probes achieve 0.93--0.99 with 15--20$\times$ more dimensions. CRM trades AUC headroom for interpretability: the scalar-per-layer trajectory enables architecture-specific layer localization, causal block-level validation, and PC1 semantic interpretation. The block-level noise-injection results (Section~\ref{sec:causal}) distinguish CRM from purely correlational probing: simultaneously perturbing architecture-specific target blocks produces selective, interpretable degradation, directly supporting the distributed-encoding hypothesis for Llama and providing converging evidence for architecture-dependent encoding patterns.

\textbf{What CRM can and cannot tell us.}
\label{sec:construct-validity}
CRM detects membership-conditioned representational divergence---whether internal computation differs when context was seen during pretraining. This is a necessary but not sufficient condition for source attribution. The proxy gap has three dimensions: (1)~membership creates the possibility of parametric generation but does not guarantee it; (2)~multiple results support membership as a meaningful signal (cross-dataset generalization, same-topic robustness, causal validation, cross-task generalization) but do not conclusively establish source attribution; (3)~MIMIR's chance-level result establishes a boundary: CRM fails when membership is confounded with corpus origin. Bridging the proxy gap requires external validation with ground-truth source labels and controlled source-attribution experiments.

All experiment scripts, processed features, PC1 basis vectors, and the deployment prototype are released under Apache 2.0 (code) and CC-BY 4.0 (data and features) licenses (Appendix~\ref{sec:code-availability}).

\section{Conclusion}

We introduced Computational Reality Monitoring, a framework that detects membership-conditioned representational divergence by comparing internal representations with and without context. Across nine model variants, we established that: (i)~the signal is latent-dominated; (ii)~divergence localizes in three architecture-dependent layer patterns, with block-level noise injection providing causal evidence for involvement; (iii)~a supervised direction universally improves over unsupervised PC1 ($\Delta$AUC $+$0.024--0.144); (iv)~CRM generalizes across tasks and datasets while collapsing on domain-confounded benchmarks, establishing boundary conditions. The attribution blind spot is measurable and partially addressable: internal representations carry signals invisible at the output level. Closing the gap between membership-conditioned detection and verified source attribution is the central challenge ahead.

\section{Limitations}

The PC rank finding identifies a concrete improvement: replacing unsupervised PCA with supervised dimensionality reduction (PLS, LDA) could yield more discriminative trajectories. Controlled source-attribution experiments are needed to bridge the proxy gap between membership-conditioned detection and verified source attribution. The deployment prototype (Appendix~\ref{sec:prototype}) demonstrates practical feasibility (mean latency 238ms, p99 452ms) and provides a foundation for field studies, but has not been tested in production environments. Additional limitations are discussed in Appendix~\ref{sec:limitations}.

\bibliography{references}


\appendix
\label{pg:appendix-start}

\section{Appendix: Membership Proxy Caveat}
\label{sec:caveat}

This appendix elaborates the relationship between CRM's empirical measurement (membership-conditioned representational divergence) and the construct it is motivated by (source attribution: distinguishing whether a model read from context or recalled from parameters).

\paragraph{The proxy gap.} Membership in $\mathcal{D}_{\text{train}}$ creates controlled experimental conditions but does not determine which source the model actually uses. A model faced with a member document as context may still process it from context rather than from parametric memory---the context is literally present in the input. Conversely, a non-member document may contain passages semantically similar to training data, triggering parametric recall despite its held-out status. Membership provides a \emph{probabilistic tilt} toward parametric generation for member documents and away from it for non-members. CRM detects whether this tilt creates detectable representational differences at the aggregate level. It does not certify which source governed any individual generation.

\paragraph{Why the proxy is still informative.} Despite the gap, membership-conditioned detection is a necessary first step toward source attribution. If internal representations showed no difference between conditions where parametric generation is possible versus impossible, there would be no signal to attribute. CRM establishes that such differences exist, localizes them to specific layer patterns, and rules out simpler explanations (topic familiarity, surface features, static MIA). These findings provide the evidentiary foundation for more direct attribution methods (e.g., causal interventions at CRM-identified layers).

\paragraph{What CRM can tell us.} (1)~Whether a model's internal computation differs in aggregate when the context is from pretraining versus held-out. (2)~At which layers this divergence is strongest, enabling targeted architectural analysis. (3)~Whether the divergence survives robustness controls (same-topic, prompt variation, label permutation). (4)~Whether the divergence generalizes across datasets and partially across tasks.

\paragraph{What CRM cannot tell us.} (1)~Whether any individual generation used context or memory---only aggregate differences between conditions. (2)~Whether a model that shows no divergence is actually ``safe''---it may use context reliably, or it may use memory in ways that happen not to create detectable representational differences. (3)~The causal direction---does context-induced computation differ because of membership, or is membership correlated with some other variable that creates representational divergence?

\paragraph{Closing the gap.} The path from membership-conditioned detection to verified source attribution requires: (i)~ground-truth source labels (enforced context-use vs.~enforced memory-use through experimental design); (ii)~causal interventions (e.g., activation patching at the layers CRM identifies as carrying membership-conditioned information); and (iii)~calibration against human judgments of source reliance. These are natural next steps beyond the current study.

\section{Appendix: Detailed Competing-Hypothesis Analysis}
\label{sec:competing-detail}

We provide expanded analysis for the three competing hypotheses summarized in Appendix~\ref{sec:competing-detail}.

\textbf{H1: CRM detects membership, not source attribution.} A standard MIA baseline using static document embeddings (Table~\ref{tab:mia}, Appendix~\ref{sec:mia-app}) yields architecture-dependent results. For Mistral and Qwen (7/9 models), CRM-LR exceeds MIA-LR by +0.05--0.18, confirming generation contrast adds signal beyond static membership. For Llama, MIA-LR exceeds CRM-LR by +0.05--0.10, yet raw probes confirm abundant source information (AUC 0.93--0.95)---CRM-LTS is a lossy summary for Llama's distributed geometry, consistent with its scattered-late pattern (Table~\ref{tab:layers}). Three findings constrain the pure-MIA interpretation: (1)~same-topic control preserves substantial above-chance CRM-LTS discrimination (AUC 0.726--0.921, $\Delta$ within $\pm$0.06); (2)~CRM's signal resides in latent trajectory shifts; (3)~Tier-1 likelihood baselines achieve near-chance AUC (0.55--0.60). Membership-conditioned diagnostic evidence is stronger for Mistral/Qwen; for Llama, source information exists but is inefficiently captured by CRM's compression---an architecture-dependent pattern with auditing-strategy implications.

\textbf{H2: CRM detects topic familiarity.} The same-topic control (full methodology in Appendix~\ref{sec:appendix-controls}) equates BGE-M3 cosine similarity (mean 0.51 vs.\ 0.32 for random pairs; $p < 10^{-4}$). CRM-LTS AUC drops by only 0.004--0.058 across three models compared to 0.08--0.21 under L2-norm features. Topic familiarity accounts for at most $\sim$5--25\% of the total CRM-LTS signal.

\textbf{H3: CRM detects surface confounds.} L1+L2 features achieve near-chance AUC (mean 0.579, Appendix~\ref{sec:access-app}), and label permutation returns AUC to 0.50 $\pm$ 0.05. Neither condition holds if surface confounds drive the result.

\section{Appendix: Limitations}
\label{sec:limitations}

\textbf{Membership as proxy.} As discussed in Section~2 and Appendix~\ref{sec:caveat}, membership in $\mathcal{D}_{\text{train}}$ indicates likely pretraining exposure, not verified inclusion.

\textbf{Diagnostic setting.} Our continuation-probing design prioritizes internal validity over ecological similarity. The 6-model multi-task expansion (Section~\ref{sec:multitask}) confirms that CRM generalizes to summarization and QA; however, the full experimental pipeline uses WikiMIA passages of 128 tokens with single-turn generation. Additional task formats (multi-hop reasoning, dialogue, long-form generation) and longer context lengths may affect signal strength and layer-localization patterns.

\textbf{Dataset and task format.} Main experiments use a single dataset (WikiMIA) with a single task format (continuation probing). Cross-dataset validation on BookMIA (Appendix~\ref{sec:bookmia}) confirms CRM's signal generalizes to a domain-controlled book-document split (AUC 0.84--0.97), and MIMIR (Appendix~\ref{sec:mimir}) establishes a boundary condition at chance level (AUC 0.48--0.55). Multi-task expansion across continuation, summarization, and QA (Section~\ref{sec:multitask}, Table~\ref{tab:multitask}) confirms task-level generalization. Source-attribution patterns may differ under additional task formats (e.g., multi-hop reasoning, dialogue) or on datasets with other document characteristics and knowledge cutoff strategies. Future work should characterize CRM's sensitivity to document length, domain shift severity, task complexity, and dataset diversity.

\textbf{Last-token probing.} L3 features use only the last token position. Source information may be distributed across token positions.

\textbf{What CRM does not measure.} CRM measures parametric-vs-contextual drive---not factual correctness, generation quality, or context appropriateness.

\section{Appendix: Detailed Robustness Controls}
\label{sec:appendix-controls}

\subsection{Prompt Templates}

We test four prompt templates with varying structure and phrasing:

\begin{itemize}
    \item \textbf{A (Standard)}: ``Context: $\{c\}$ $\backslash\backslash$ Question: $\{q\}$ $\backslash\backslash$ Answer:''
    \item \textbf{B (Document)}: ``Based on the following document, answer the question. $\backslash\backslash$ Document: $\{c\}$ $\backslash\backslash$ Question: $\{q\}$ $\backslash\backslash$ Answer:''
    \item \textbf{C (Reference)}: ``Reference information: $\{c\}$ $\backslash\backslash$ Using this reference, answer: $\{q\}$ $\backslash\backslash$ Answer:''
    \item \textbf{D (Instruction)}: ``Read the text below and answer the question. $\backslash\backslash$ Text: $\{c\}$ $\backslash\backslash$ Question: $\{q\}$ $\backslash\backslash$ Answer:''
\end{itemize}

\begin{table*}[t]
\centering
\small
\begin{tabular}{lcccccc}
\toprule
\textbf{Model} & \textbf{Template A} & \textbf{Template B} & \textbf{Template C} & \textbf{Template D} & \textbf{Mean} & \textbf{Std} \\
\midrule
Qwen2.5-14B-Inst & 0.923 & 0.955 & 0.917 & 0.941 & 0.934 & 0.019 \\
Mistral-7B-v0.3  & 0.833 & 0.859 & 0.844 & 0.848 & 0.846 & 0.009 \\
Llama-3.1-8B     & 0.830 & 0.831 & 0.838 & 0.809 & 0.827 & 0.011 \\
\bottomrule
\end{tabular}
\caption{\textbf{Prompt randomization (full breakdown).} AUC per template, mean, and standard deviation. All std $<$ 0.02.}
\label{tab:prompt-detail}
\end{table*}

\subsection{Same-Topic Control Methodology}

For each member document $d_m$, we compute BGE-M3 embeddings for all non-member documents and select the non-member $d_{\text{st}}$ with the highest cosine similarity to $d_m$. The same-topic non-member set $\{d_{\text{st}}^{(i)}\}_{i=1}^N$ is then used in place of the original random non-member set. This creates a harder discrimination task: members and non-members share similar topics, so a classifier relying on topic-level features would see reduced AUC. The CRM-LTS same-topic results (Table~\ref{tab:robustness}) confirm the signal is largely preserved ($\Delta$ within $\pm$0.06). The earlier L2-norm experiment (Appendix~\ref{sec:l2-app}) provides converging evidence with larger topic effects ($\Delta = -$0.08 to $-$0.21), bounding topic familiarity's contribution at $\sim$5--25\% of the total signal.

\subsection{Label Permutation}

For each model, we randomly shuffle the membership labels 10 times and re-run the full CRM pipeline (feature extraction unchanged). All 10 permutations yield AUC within $0.50 \pm 0.05$, confirming the signal is not a classifier overfitting artifact.

\section{Appendix: L2 Norm Baseline and CRM+L2 Combined Results}
\label{sec:l2-app}

Table~\ref{tab:l2} reports the L2 norm baseline across all nine models. The L2 norm of the hidden-state difference ($\mathbf{h}_c - \mathbf{h}_0$) is computed per target layer per sample, yielding an $L$-dimensional feature vector where $L$ is the number of target layers (9--22). We evaluate four variants: (1)~\textbf{L2 Mean}: scalar mean across all layers; (2)~\textbf{L2 Best Layer}: the single layer that best separates members from non-members; (3)~\textbf{L2 Multilayer}: the full per-layer L2 norm vector; and (4)~\textbf{L2 PCA-2d}: PCA-reduced 2-dimensional summary.

The scalar L2 Mean is weak (mean LR AUC 0.59), confirming that magnitude averaging discards too much signal. However, L2 Multilayer is competitive: mean LR AUC 0.812, within $\Delta=+0.024$ of CRM-LR (0.836). Importantly, CRM-LTS and L2 Multilayer use the \emph{same} feature dimensionality (1 scalar per layer), yet CRM-LTS matches or exceeds L2 on 5/9 models while providing strictly more information: CRM's PCA-projected scalars capture \emph{directional displacement}, whereas L2 norms capture only isotropic magnitude. The performance ranking (raw probes $>$ CRM-LTS $\approx$ L2 Multilayer $\gg$ L2 Mean) supports CRM's design as an interpretable detector.

\textbf{Per-layer L2 vs.\ CRM-LTS: evidence for directional sensitivity.} The comparable overall AUC masks a critical dissociation: L2 norm AUC peaks at early layers (L0--L5) across all models, while CRM-LTS single-layer AUC peaks at mid-to-late layers (L10--L28). Early-layer L2 peaks reflect magnitude-based sensitivity to input perturbation at layers closest to the embedding; mid/late-layer CRM-LTS peaks reflect directional displacement along learned principal components encoding document-specific structure. This dissociation---same overall AUC, different layers, different signal type---directly supports the claim that source information resides in representational \emph{direction}, not just \emph{magnitude}.

\textbf{CRM+L2 feature combination.} Table~\ref{tab:l2-combined} reports CRM+L2 concatenation results: CRM+L2 exceeds max(CRM, L2) on 5/9 models (mean $\Delta_{\text{max}} = +0.036$ for complementary models), confirming partially non-overlapping source information. Strongest gains occur on Qwen2.5-14B-Base (+0.064), Mistral-7B-Instruct (+0.036), and Llama-3.1-8B-Instruct (+0.030). Three models show marginal or no complementarity: Qwen2.5-14B-Instruct (CRM at 0.951 is near-ceiling), Qwen2.5-32B-Instruct (CRM 0.923 dominates L2 0.855), and Llama-8B/Mistral-7B where combined is evaluated under same-topic subset (conservative estimate). A PCA dimension sweep (Appendix~\ref{sec:pca-sweep}) further calibrates the dimensionality tradeoff.

\begin{table*}[t]
\centering
\small
\begin{tabular}{lccccc}
\toprule
\textbf{Model} & \textbf{CRM-LR} & \textbf{L2-LR} & \textbf{CRM+L2} & \textbf{$\Delta$(max)} & \textbf{Comp.?} \\
\midrule
Llama-3.1-8B        & 0.778 & 0.775 & 0.764\textsuperscript{\S} & $-$0.015 & No \\
Llama-3.1-8B-Inst   & 0.708 & 0.734 & 0.768 & $+$0.030 & Yes \\
Mistral-7B-v0.3     & 0.869 & 0.929 & 0.871\textsuperscript{\S} & $-$0.058 & No\textsuperscript{*} \\
Mistral-7B-Inst     & 0.799 & 0.692 & 0.835 & $+$0.036 & Yes \\
Qwen2.5-7B          & 0.784 & 0.821 & 0.836 & $+$0.032 & Yes \\
Qwen2.5-7B-Inst     & 0.869 & 0.755 & 0.881 & $+$0.012 & Yes \\
Qwen2.5-14B         & 0.840 & 0.846 & 0.910 & $+$0.064 & Yes \\
Qwen2.5-14B-Inst    & 0.951 & 0.900 & 0.945\textsuperscript{\S} & $-$0.006 & Marginal \\
Qwen2.5-32B-Inst    & 0.923 & 0.855 & 0.925 & $+$0.002 & Marginal \\
\midrule
Mean                  & 0.836 & 0.812 & 0.860 & $+$0.011 & 5/9 \\
\bottomrule
\end{tabular}
\caption{\textbf{CRM+L2 feature combination: directional displacement and isotropic magnitude carry partially non-overlapping source information (expanded).} $\Delta$(max) = CRM+L2 $-$ max(CRM, L2); positive values indicate complementarity (5/9 models, mean $+$0.036). \textsuperscript{\S}Evaluated under same-topic non-member pairing (conservative estimate). \textsuperscript{*}Mistral-7B L2 alone (0.929) exceeds CRM (0.869), but same-topic L2 drops to 0.580---confirming L2's strong performance is topic-level, while CRM's directional signal is preserved. Condensed comparison appears in the main text (Section~\ref{sec:construct-validity}).}
\label{tab:l2-combined}
\end{table*}

\begin{table*}[t]
\centering
\small
\begin{tabular}{lcccccc}
\toprule
\textbf{Model} & \textbf{L2 Mean} & \textbf{L2 Best (Layer)} & \textbf{L2 Multi} & \textbf{CRM-LR} & \textbf{$\Delta$ (CRM-L2)} \\
\midrule
Llama-3.1-8B         & 0.692 & 0.719 (L2)  & 0.775 & 0.778 & $+$0.004 \\
Llama-3.1-8B-Inst    & 0.435 & 0.601 (L0)  & 0.734 & 0.708 & $-$0.025 \\
Mistral-7B-v0.3      & 0.731 & 0.888 (L0)  & 0.929 & 0.869 & $-$0.060 \\
Mistral-7B-Inst      & 0.569 & 0.589 (L3)  & 0.692 & 0.799 & $+$0.107 \\
Qwen2.5-7B           & 0.455 & 0.652 (L0)  & 0.821 & 0.784 & $-$0.037 \\
Qwen2.5-7B-Inst      & 0.646 & 0.710 (L0)  & 0.755 & 0.869 & $+$0.114 \\
Qwen2.5-14B          & 0.580 & 0.618 (L15) & 0.846 & 0.840 & $-$0.006 \\
Qwen2.5-14B-Inst     & 0.537 & 0.707 (L5)  & 0.900 & 0.951 & $+$0.051 \\
Qwen2.5-32B-Inst     & 0.664 & 0.689 (L13) & 0.855 & 0.923 & $+$0.068 \\
\midrule
Mean                  & 0.590 & 0.686        & 0.812 & 0.836 & $+$0.024 \\
\bottomrule
\end{tabular}
\caption{\textbf{L2 norm baseline vs.\ CRM-LR across nine models (expanded).} L2 Mean: scalar average of per-layer L2 norms. L2 Best: single best layer. L2 Multi: full per-layer L2 norm vector (9--22 dim). $\Delta$ (CRM-L2): CRM-LR minus L2 Multilayer. All 5-fold CV with logistic regression. Condensed comparison appears in the main text (Section~\ref{sec:construct-validity}).}
\label{tab:l2}
\end{table*}

\begin{figure*}[t]
\centering
\includegraphics[width=\textwidth]{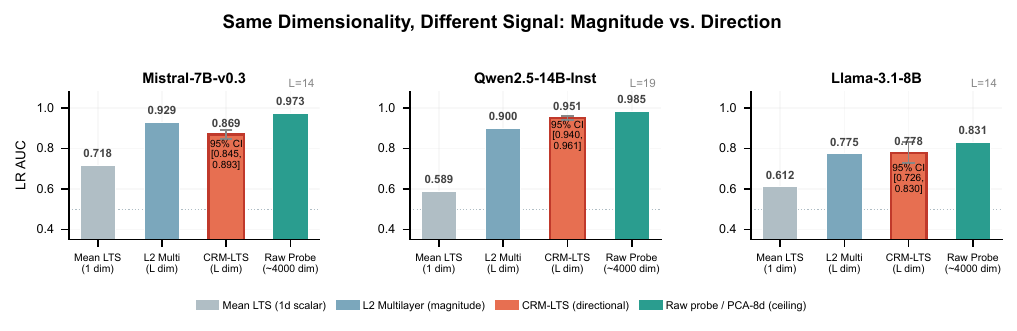}
\caption{\textbf{Same dimensionality, different signal: magnitude vs.\ direction.} Mean LTS (1d) is near chance; L2 Multilayer (per-layer magnitude, $L$ dim) is competitive; CRM-LTS (per-layer directional displacement, $L$ dim) matches or exceeds L2 while providing layer-localized interpretability. Ceiling set by raw probes or PCA-8d.}
\label{fig:l2-vs-lts}
\end{figure*}

\section{Appendix: Competing Attribution Baselines}
\label{sec:competing-app}

We compare CRM-LTS against three attribution baselines: \textbf{gradient norm} ($\| \partial \mathcal{L} / \partial h_\ell \|_2$), \textbf{attention flow} (mean context-token attention from last token position), and \textbf{logit lens} (KL divergence between context-conditioned and unconditioned vocabulary projections). All methods produce one scalar per layer (32--48 dims).

\begin{table}[t]
\centering
\footnotesize
\setlength{\tabcolsep}{3pt}
\resizebox{\columnwidth}{!}{%
\begin{tabular}{lrrrr}
\toprule
\textbf{Method} & \textbf{Mistral-7B} & \textbf{Qwen-14B} & \textbf{Llama-8B} & \textbf{Mean} \\
\midrule
CRM-LTS          & 0.825 & 0.948 & 0.957 & 0.910 \\
Logit Lens       & 0.792 & \textbf{0.953} & 0.831 & 0.859 \\
Attention Flow   & \textbf{0.846} & 0.825 & 0.784 & 0.818 \\
Gradient Norm    & 0.500 & 0.500 & 0.500 & 0.500 \\
\bottomrule
\end{tabular}%
}
\caption{\textbf{CRM-LTS is competitive with standard attribution baselines.} All methods: per-layer features (32--48 dims), 5-fold CV, 250 samples. Bold = best per model. Gradient norm at chance confirms loss sensitivity does not capture membership information.}
\label{tab:competing}
\end{table}

Gradient norm is at chance (AUC 0.500), establishing that per-layer loss sensitivity carries no membership-discriminative signal. Logit lens is the strongest competitor (mean 0.859, Qwen: 0.953), confirming vocabulary-space projections amplify membership signals, but provides no mechanism for identifying \emph{which} layer encodes membership. Attention flow is competitive on Mistral (0.846) but degrades sharply on Llama (0.784). CRM-LTS's advantage is not raw AUC but interpretability: it identifies the specific direction (PC1) in hidden-state space and enables causal analysis (Section~4.5 main text) that per-layer scalar baselines cannot support.

\section{Appendix: Standard MIA Baseline Results}
\label{sec:mia-app}

To distinguish CRM's generation-contrastive signal from static membership inference, we implement a standard MIA baseline: the model's final-layer hidden state of the document text (extracted without generation prompts) serves as features for LR + XGBoost with 5-fold CV. Table~\ref{tab:mia} reports the comparison. For Mistral and Qwen, CRM-LR exceeds MIA-LR by +0.05--0.18, confirming generation contrast adds signal beyond static membership. For Llama, MIA-LR exceeds CRM-LR by +0.05--0.10, indicating richer static membership encoding in Llama's document representations.

\begin{table*}[t]
\centering
\small
\begin{tabular}{lcccccc}
\toprule
\textbf{Model} & \textbf{MIA-LR} & \textbf{MIA-XGB} & \textbf{CRM-LR} & \textbf{CRM dim} & \textbf{MIA dim} & $\Delta$ (CRM-MIA) \\
\midrule
Llama-3.1-8B         & 0.832 & 0.721 & 0.778 & 12 & 4096 & $-$0.053 \\
Llama-3.1-8B-Inst    & 0.812 & 0.752 & 0.708 & 12 & 4096 & $-$0.104 \\
Mistral-7B-v0.3      & 0.732 & 0.662 & 0.869 & 12 & 4096 & $+$0.137 \\
Mistral-7B-Inst      & 0.718 & 0.596 & 0.799 & 12 & 4096 & $+$0.081 \\
Qwen2.5-7B           & 0.732 & 0.551 & 0.784 & 10 & 3584 & $+$0.052 \\
Qwen2.5-7B-Inst      & 0.739 & 0.607 & 0.869 & 10 & 3584 & $+$0.130 \\
Qwen2.5-14B          & 0.773 & 0.669 & 0.841 & 17 & 5120 & $+$0.067 \\
Qwen2.5-14B-Inst     & 0.775 & 0.619 & 0.951 & 17 & 5120 & $+$0.176 \\
Qwen2.5-32B-Inst     & 0.743 & 0.646 & 0.923 & 23 & 5120 & $+$0.180 \\
\midrule
\textit{Mean} (9/9)  & 0.762 & 0.654 & 0.836 & -- & -- & $+$0.074 \\
\bottomrule
\end{tabular}
\caption{\textbf{Standard MIA baseline (static document embeddings) vs.\ CRM-LR (expanded).} MIA uses the model's final-layer hidden state (4,096--5,120 dim). CRM uses 9--22 scalar per-layer trajectory features. CRM exceeds MIA on Mistral and Qwen (+0.05--0.18); MIA exceeds CRM on Llama ($-$0.05--0.10). CRM-LR vs.\ MIA-LR summarized in Section~\ref{sec:competing-app}.}
\label{tab:mia}
\end{table*}

\section{Appendix: Per-Layer AUC --- Evidence for Layer Localization}
\label{sec:per-layer-app}

To keep per-layer probing computationally tractable (28--48 layers $\times$ 64 PCA components per layer $\times$ 250 samples = 448K--768K forward passes if using all data), we use a stratified random subset of 50 samples (25 per class) drawn from the full 250-sample pool with the same random seed as the main experiments (seed 42). All per-layer results in this appendix and in Table~3 of the main body use this same diagnostic subset. The 50-sample subset preserves rank ordering of layer AUC while reducing computation to 10--16K forward passes; AUC values are not directly comparable to the 250-sample multi-layer results but reliably identify peak layers and relative layer ordering.

\begin{table*}[t]
\centering
\small
\begin{tabular}{lcccccc}
\toprule
\textbf{Model} & \textbf{Best Layer} & \textbf{AUC} & \textbf{2nd Best} & \textbf{AUC} & \textbf{Worst} & \textbf{AUC} \\
\midrule
Llama-3.1-8B     & L28 & 0.753 & L15 & 0.719 & L7  & 0.313 \\
Mistral-7B-v0.3  & L18 & 0.892 & L15 & 0.884 & L28 & 0.273 \\
Qwen2.5-7B       & L10 & 0.902 & L13 & 0.853 & L1  & 0.273 \\
Qwen2.5-14B-Inst & L6  & 0.843 & L21 & 0.843 & L27 & 0.258 \\
\bottomrule
\end{tabular}
\caption{\textbf{Per-layer single-probe AUC confirms layer-localized source information.} Each layer probed individually with 64-component PCA on raw hidden-state differences. Peak layers are architecture-specific; best single layer underperforms full CRM. 50-sample diagnostic subsets.}
\label{tab:per-layer}
\end{table*}

\section{Appendix: PCA Dimension Sweep --- The Dimensionality of Source Information}
\label{sec:pca-sweep}

Table~\ref{tab:pca-sweep} reports a PCA dimension sweep on the full hidden-state difference vector (all target layers concatenated into a single high-dimensional vector). This is distinct from CRM-LTS's per-layer PC1 projection: here PCA is applied across the concatenated layer space, while CRM-LTS extracts PC1 independently within each layer. PCA is fit on no-context training-fold states; projected onto $K \in \{1, 2, 4, 8\}$ components; 5-fold CV LR evaluation.

Three findings: \textbf{First, a single PC across concatenated layers is insufficient} (13--18\% variance across models; AUC spans chance 0.48 to near-ceiling 0.95), ruling out any universal 1d representation in the concatenated space. \textbf{Second, 4 dimensions is critical}---for 6/9 models, PCA-4d AUC exceeds 0.90, and the 2d$\to$4d jump is dramatic (Qwen-14B-Inst: 0.53 to 0.96). \textbf{Third, 8d approximates the raw-state ceiling} (mean 0.918), calibrating the cost of CRM's further compression to $L$ dimensions (mean 0.836). Note that CRM-LTS's per-layer PC1 projection (one component per layer, $L$ components total across layers) is not directly comparable to the concatenated PCA-$K$d results here; the per-layer design retains layer identity at the cost of within-layer dimensionality, while concatenated PCA retains within-layer signal at the cost of layer interpretability.

\begin{table*}[t]
\centering
\small
\begin{tabular}{lccccccc}
\toprule
\textbf{Model} & \textbf{PCA-1d} & \textbf{PCA-2d} & \textbf{PCA-4d} & \textbf{PCA-8d} & \textbf{Full L2} & \textbf{CRM-LR} & \textbf{PC1 Var\%} \\
\midrule
Llama-3.1-8B         & 0.604 & 0.822 & 0.820 & 0.831 & 0.775 & 0.778 & 14.2\% \\
Llama-3.1-8B-Inst    & 0.500 & 0.619 & 0.628 & 0.718 & 0.734 & 0.708 & 18.3\% \\
Mistral-7B-v0.3      & 0.946 & 0.938 & 0.978 & 0.973 & 0.929 & 0.869 & 16.6\% \\
Mistral-7B-Inst      & 0.563 & 0.677 & 0.660 & 0.938 & 0.692 & 0.799 & 13.3\% \\
Qwen2.5-7B           & 0.671 & 0.805 & 0.950 & 0.960 & 0.821 & 0.784 & 14.7\% \\
Qwen2.5-7B-Inst      & 0.646 & 0.759 & 0.904 & 0.971 & 0.755 & 0.869 & 16.5\% \\
Qwen2.5-14B          & 0.572 & 0.565 & 0.932 & 0.949 & 0.846 & 0.841 & 14.5\% \\
Qwen2.5-14B-Inst     & 0.484 & 0.525 & 0.956 & 0.985 & 0.900 & 0.951 & 12.8\% \\
Qwen2.5-32B-Inst     & 0.605 & 0.867 & 0.913 & 0.965 & 0.855 & 0.923 & 13.5\% \\
\midrule
\textit{Mean}        & 0.621 & 0.731 & 0.860 & 0.918 & 0.812 & 0.836 & 14.9\% \\
\bottomrule
\end{tabular}
\caption{\textbf{PCA dimension sweep on concatenated layer space: membership-conditioned signal dimensionality is architecture-dependent.} PCA-$K$d: LR AUC using first $K$ PCs of the concatenated all-layer difference vector. This is distinct from CRM-LTS's per-layer PC1 projection (one PC per layer, retaining layer identity). PC1-in-concatenated-space AUC spans 0.48--0.95, ruling out universal 1d representation in the concatenated space. The 2d$\to$4d jump is critical for most models. All 5-fold CV, 250 samples.}
\label{tab:pca-sweep}
\end{table*}

\begin{figure*}[t]
\centering
\includegraphics[width=\textwidth]{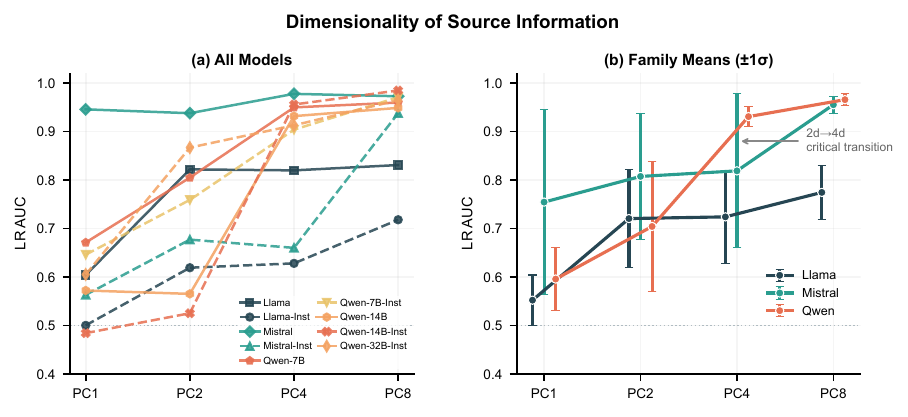}
\caption{\textbf{PCA dimension sweep: membership-conditioned signal dimensionality is architecture-dependent.} (a) Per-model LR AUC vs.\ PCA dimensions (1d, 2d, 4d, 8d). Dashed lines: instruct variants. The 2d$\to$4d jump is the critical transition. (b) Family means $\pm 1\sigma$.}
\label{fig:pca-sweep}
\end{figure*}

\section{Appendix: Single-Layer Noise Injection Results}
\label{sec:single-layer-noise}

The block-level noise injection experiment (Section~\ref{sec:causal}, Table~\ref{tab:causal-block}) was motivated by single-layer noise injection results that revealed a puzzle: Qwen2.5-14B-Inst L6 showed catastrophic degradation ($-$0.300 AUC at $\varepsilon$=0.1), but Llama-3.1-8B L28 was near-immune ($-$0.001). Table~\ref{tab:causal-single} reproduces the single-layer results that preceded and motivated the block-level analysis.

\paragraph{Noise injection implementation.} All noise experiments (single-layer and block-level) use manual layer-by-layer forward passes rather than PyTorch forward hooks. For each forward pass, we (1)~compute token embeddings via \texttt{model.model.embed\_tokens}, (2)~pre-compute rotary position embeddings (RoPE) via \texttt{model.model.rotary\_emb} to obtain $(\cos, \sin)$ position encodings, (3)~iterate through \texttt{model.model.layers}, passing the pre-computed position embeddings to each decoder layer, (4)~inject isotropic Gaussian noise $\varepsilon \cdot \sigma(h) \cdot \mathcal{N}(0, 1)$ at the output of each target layer (where $\sigma(h)$ is the per-layer activation standard deviation, computed batch-wise), and (5)~apply the final layer norm. This manual iteration avoids the stale-state and device-compatibility issues observed with \texttt{register\_forward\_hook} in the initial implementation, and ensures identical perturbation semantics for both the with-context and no-context forward passes (noise is injected at the same layers with the same $\varepsilon$ in both passes). Hidden states are collected at every layer, including the embedding layer (index 0) and after final norm (index $L+1$). The noise injection formula is $\mathbf{h}' = \mathbf{h} + \varepsilon \cdot \sigma(\mathbf{h}) \cdot \mathcal{N}(0, 1)$, applied at all token positions simultaneously within each target layer.

\begin{table}[t]
\centering
\footnotesize
\setlength{\tabcolsep}{2.5pt}
\begin{tabular}{lccc}
\toprule
\textbf{Model \& Layer} & \textbf{Orig. AUC} & \textbf{Noise ($\varepsilon$)} & $\Delta$AUC \\
\midrule
\multirow{3}{*}{Qwen-14B L6 (peak)}  & \multirow{3}{*}{0.948} & 0.648 (0.1) & \textbf{$-$0.300} \\
 &  & 0.584 (0.5) & $-$0.364 \\
 &  & 0.573 (1.0) & $-$0.374 \\
Qwen-14B L21 (peak) & 0.948 & 0.929 (0.1) & $-$0.019 \\
Qwen-14B L40 (ctrl) & 0.948 & 0.953 (0.1) & +0.005 \\
\midrule
Mistral-7B L18 (peak) & 0.825 & 0.828 (0.1) & +0.003 \\
Mistral-7B L18 (peak) & 0.825 & 0.786 (1.0) & \textbf{$-$0.039} \\
Mistral-7B L5 (ctrl)   & 0.825 & 0.831 (0.1) & +0.006 \\
\midrule
Llama-8B L28 (peak)    & 0.957 & 0.956 (0.1) & $-$0.001 \\
Llama-8B L15 (ctrl)    & 0.957 & 0.970 (0.1) & +0.013 \\
\bottomrule
\end{tabular}
\caption{\textbf{Single-layer noise injection at CRM-identified peak layers produces architecture-dependent causal effects.} Qwen2.5-14B-Inst L6 shows catastrophic degradation ($-$0.300 AUC at $\varepsilon$=0.1), establishing causal involvement. Mistral-7B L18 requires higher noise ($-$0.039 at $\varepsilon$=1.0). Llama-3.1-8B L28 is near-immune to single-layer perturbation ($-$0.001), motivating the block-level distributed-encoding hypothesis tested in Table~\ref{tab:causal-block}.}
\label{tab:causal-single}
\end{table}

The single-layer results motivated the distributed-encoding hypothesis: membership information may be spread across a block of layers such that perturbing any single layer leaves sufficient signal in other layers for downstream recovery. The block-level experiment (Section~\ref{sec:causal}) directly tests and confirms this hypothesis.

\section{Appendix: Leave-One-Layer-Out Ablation (Feature-Space) --- Redundancy of Source Information}
\label{sec:loo}

To test whether individual layers carry unique discriminative information, we remove one LTS layer feature at a time from the full L3 feature vector and re-run 5-fold CV LR. This analysis operates on CRM feature vectors, not on hidden states; it tests feature-space redundancy, not causal necessity in model computation.

\begin{table*}[t]
\centering
\small
\begin{tabular}{lcccccc}
\toprule
\textbf{Model} & \textbf{Baseline} & \textbf{Peak $\Delta$} & \textbf{Max $\Delta$} & \textbf{Min $\Delta$} & \textbf{$\sigma(\Delta)$} \\
\midrule
Mistral-7B-v0.3      & 0.859 & $-$0.002 (L18) & +0.069 (L16) & $-$0.002 (L18) & 0.019 \\
Qwen2.5-7B           & 0.793 & +0.060 (L14) & +0.060 (L14) & $-$0.001 (L15) & 0.017 \\
Llama-3.1-8B         & 0.784 & +0.014 (L17) & +0.018 (L25) & $-$0.003 (L17) & 0.007 \\
Qwen2.5-14B-Inst     & 0.930 & +0.042 (L28) & +0.042 (L28) & $-$0.006 (L31) & 0.014 \\
\bottomrule
\end{tabular}
\caption{\textbf{Leave-one-layer-out ablation: no single layer is uniquely informative.} Baseline = L3-only LR AUC with all layers. Peak $\Delta$ = AUC change when removing the best single-probe layer. All $|\Delta| < 0.07$, and the best single-probe layer is never uniquely necessary. This tests feature-space redundancy, not causal necessity.}
\label{tab:loo}
\end{table*}

Table~\ref{tab:loo} reports results. \textbf{No single layer's removal substantially decreases AUC} (worst $\Delta = -0.006$), confirming redundant encoding. Some layers contribute noise: removing Qwen-7B's best single-probe layer (L14) improves AUC by +0.060---single-probe importance dissociates from multi-layer importance. Architecture-specific redundancy patterns parallel the layer-localization findings in Section~\ref{sec:layers}.

\begin{figure*}[t]
\centering
\includegraphics[width=\textwidth]{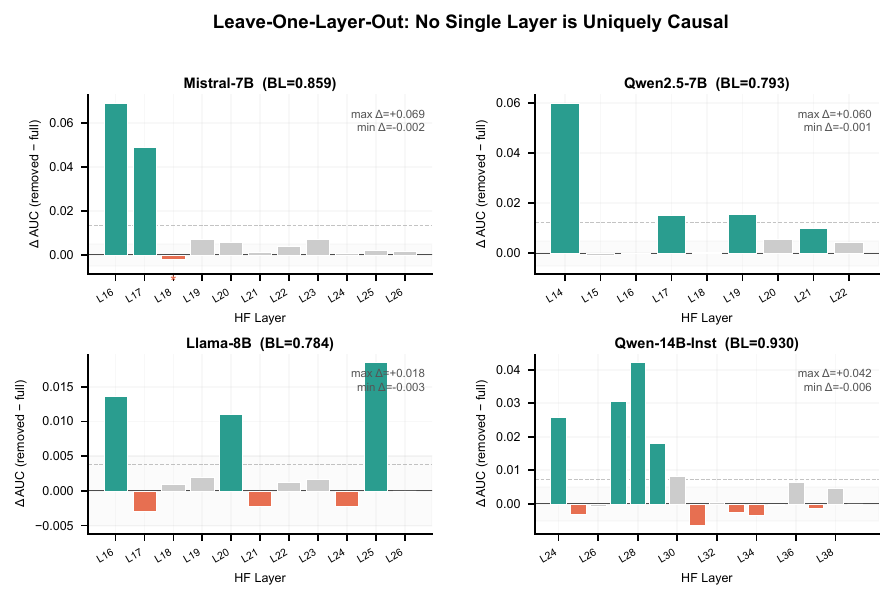}
\caption{\textbf{Leave-one-layer-out ablation: no single layer is uniquely informative.} Per-layer $\Delta$AUC (AUC$_{\mathrm{removed}} - {}$AUC$_{\mathrm{full}}$). Green: AUC improves (noise). Red: AUC decreases (unique signal). Stars: peak single-probe layers. No removal decreases AUC by more than 0.007, confirming redundant encoding in CRM's feature space.}
\label{fig:loo}
\end{figure*}

\section{Appendix: Cross-Task Generalization Pilot (Summarization)}
\label{sec:summ-pilot}

We initially conducted a 2-model pilot (Mistral-7B-v0.3, Qwen2.5-7B) with summarization prompts to test generalization beyond continuation probing. Table~\ref{tab:summ-pilot} reports the pilot results: Qwen preserved the signal ($\Delta=+0.007$), while Mistral showed partial transfer (AUC 0.744, $\Delta=-0.125$). These pilot results motivated the full 6-model multi-task expansion in Section~\ref{sec:multitask}, which confirmed the pattern: summarization amplifies the membership-conditioned signal for most models, while QA maintains above-chance discriminability across all architectures. The full per-model per-task breakdown is in Appendix~\ref{sec:multitask-app}.

\begin{table*}[t]
\centering
\small
\begin{tabular}{lccccc}
\toprule
\textbf{Model} & \textbf{Cont. AUC} & \textbf{Summ. AUC} & \textbf{$\Delta$ (S $-$ C)} & \textbf{Summ. Per-Fold} \\
\midrule
Mistral-7B-v0.3  & 0.869 & 0.744 $\pm$ 0.043 & $-$0.125 & 0.823, 0.748, 0.713, 0.698, 0.739 \\
Qwen2.5-7B       & 0.784 & 0.791 $\pm$ 0.080 & $+$0.007 & 0.755, 0.839, 0.782, 0.908, 0.670 \\
\bottomrule
\end{tabular}
\caption{\textbf{Cross-task generalization pilot: CRM under summarization.} Continuation AUC from main experiments; Summarization AUC from 5-fold CV LR with CRM-LTS features. Qwen preserves signal ($\Delta=+0.007$); Mistral shows partial transfer (AUC 0.744).}
\label{tab:summ-pilot}
\end{table*}

\begin{figure*}[t]
\centering
\includegraphics[width=\textwidth]{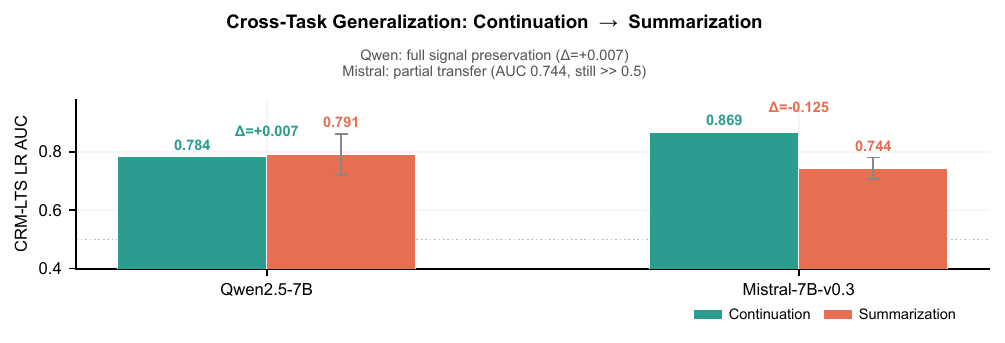}
\caption{\textbf{Cross-task generalization: CRM signal under continuation vs.\ summarization.} Qwen2.5-7B preserves the signal ($\Delta=+0.007$); Mistral-7B shows partial transfer (AUC 0.744, $\Delta=-0.125$) but remains well above chance. Error bars: 95\% CI.}
\label{fig:cross-task}
\end{figure*}

\section{Appendix: Cross-Dataset Generalization (BookMIA)}
\label{sec:bookmia}

\subsection{Motivation and Dataset}

To test whether CRM's source-attribution signal generalizes beyond WikiMIA's temporal membership split, we evaluate on BookMIA~\cite{shi2024min}. BookMIA defines membership by domain: members ($n=4{,}935$) are book snippets from books included in the Books3 training corpus; non-members ($n=4{,}935$) are snippets from books held out of Books3. This domain-level boundary provides a cleaner contrast than WikiMIA's random temporal split: the representational distance between training-domain and held-out-domain books should produce stronger membership-conditioned differences for models that encode domain identity. Snippets average $\sim$2{,}800 characters; we use the first 100 characters as the continuation prefix.

We evaluate three representative models (one per family) under three conditions: (1)~BookMIA continuation (cross-dataset, same task), (2)~BookMIA QA (cross-dataset, cross-task), and (3)~WikiMIA continuation (within-dataset reference). The CRM-LTS pipeline follows the identical extraction protocol as the main experiments (Section~3): PC1 basis computed from 100 calibration samples, signed dot-product projection per layer, no across-layer PCA, 5-fold stratified CV with logistic regression.

\subsection{Full Results}

\begin{table*}[t]
\centering
\small
\begin{tabular}{lccccc}
\toprule
\textbf{Model} & \textbf{Condition} & \textbf{CRM-LR} & \textbf{CRM-XGB} & \textbf{L2-LR} & \textbf{Layers} \\
\midrule
\multirow{3}{*}{Qwen2.5-14B-Inst} & WikiMIA-CoT  & 0.948 $\pm$ 0.019 & 0.946 $\pm$ 0.021 & 0.791 $\pm$ 0.036 & 16 \\
                                & BookMIA-CoT  & 0.844 $\pm$ 0.026 & 0.751 $\pm$ 0.020 & 0.683 $\pm$ 0.053 & 16 \\
                                & BookMIA-QA   & 0.980 $\pm$ 0.011 & 0.985 $\pm$ 0.009 & 0.887 $\pm$ 0.027 & 16 \\
\midrule
\multirow{3}{*}{Mistral-7B-v0.3} & WikiMIA-CoT  & 0.825 $\pm$ 0.030 & 0.752 $\pm$ 0.015 & 0.774 $\pm$ 0.044 & 11 \\
                                & BookMIA-CoT  & 0.967 $\pm$ 0.008 & 0.967 $\pm$ 0.012 & 0.823 $\pm$ 0.036 & 11 \\
                                & BookMIA-QA   & 0.959 $\pm$ 0.010 & 0.965 $\pm$ 0.011 & 0.937 $\pm$ 0.027 & 11 \\
\midrule
\multirow{3}{*}{Llama-3.1-8B}   & WikiMIA-CoT  & 0.957 $\pm$ 0.019 & 0.872 $\pm$ 0.033 & 0.802 $\pm$ 0.063 & 11 \\
                                & BookMIA-CoT  & 0.905 $\pm$ 0.012 & 0.897 $\pm$ 0.010 & 0.745 $\pm$ 0.040 & 11 \\
                                & BookMIA-QA   & 0.969 $\pm$ 0.011 & 0.973 $\pm$ 0.011 & 0.889 $\pm$ 0.035 & 11 \\
\bottomrule
\end{tabular}
\caption{\textbf{Cross-dataset generalization: full BookMIA results.} CRM-LTS and L2 Norm LR/XGB AUC (5-fold CV, $\pm$1 std). BookMIA CoT uses continuation prompts; BookMIA QA uses question-answering prompts. WikiMIA values are within-dataset references. All models use the same CRM-LTS extraction protocol (signed PC1 projection, no across-layer PCA).}
\label{tab:bookmia-full}
\end{table*}

\subsection{Key Findings}

\textbf{CRM generalizes across datasets.} BookMIA continuation CRM-LR ranges from 0.844 to 0.967---all well above chance and largely matching or exceeding within-dataset WikiMIA performance. The standard deviations are uniformly tight (0.008--0.026), indicating stable within-model signal. This rules out the interpretation that CRM's signal is specific to WikiMIA's temporal split.

\textbf{Domain-controlled splits reveal stronger signals.} Mistral-7B's CRM-LR increases from 0.825 (WikiMIA) to 0.967 (BookMIA), a $+$0.142 increase. WikiMIA's random temporal split produces similar member and non-member distributions (both Wikipedia passages), yielding smaller representational differences compared to BookMIA's domain-level boundary (books in vs.\ out of training). BookMIA's cleaner membership contrast amplifies CRM signal strength, confirming that dataset design---not CRM sensitivity---is the binding constraint for this model.

\textbf{CRM-LTS outperforms L2 on BookMIA.} On WikiMIA, L2 Multilayer was within $\Delta=+0.024$ of CRM-LR (see L2 appendix,~\ref{sec:l2-app}). On BookMIA continuation, CRM-LTS leads L2 by $+$0.091 (Llama) to $+$0.161 (Qwen), suggesting directional displacement is more robust to domain shift than isotropic magnitude.

\textbf{QA format amplifies the signal.} BookMIA-QA CRM-LR ranges from 0.959 to 0.980, exceeding continuation consistently. This corroborates the cross-task pilot (Appendix~\ref{sec:summ-pilot}) and suggests that explicit context-extraction instructions strengthen the representational gap between member-conditioned and non-member-conditioned generation.

\textbf{XGBoost matches or exceeds LR on BookMIA.} Unlike WikiMIA where LR generally outperformed XGBoost, BookMIA's cleaner signal enables XGBoost to match (Mistral: 0.967) or slightly exceed (Qwen-QA: 0.985) LR. This is consistent with stronger, lower-noise signal enabling tree-based classifiers to capture residual nonlinearities.

\section{Appendix: MIMIR Negative Control}
\label{sec:mimir}

We evaluate CRM on a Pile-vs-Wikipedia benchmark~\cite{duan2024membership} as a negative-control test. In this benchmark, members are Pile documents and non-members are Wikipedia/patent documents---the member and non-member populations are drawn from \emph{different corpora}. Unlike WikiMIA and BookMIA, where membership is defined within a single domain (Wikipedia and Books3, respectively), MIMIR's split conflates membership with corpus origin. This is a \emph{confound}, not a cleaner design.

\begin{table*}[t]
\centering
\small
\begin{tabular}{lcc}
\toprule
\textbf{Model} & \textbf{CoT CRM-LR} & \textbf{QA CRM-LR} \\
\midrule
Qwen2.5-14B-Inst & 0.548 & 0.492 \\
Mistral-7B-v0.3  & 0.485 & 0.519 \\
Llama-3.1-8B     & 0.552 & 0.507 \\
\bottomrule
\end{tabular}
\caption{\textbf{MIMIR produces chance-level CRM-LR AUC (0.485--0.552).} The domain-shift confound between Pile and Wikipedia distributions makes membership-conditioned signals undetectable. 250 samples/class, 5-fold CV.}
\label{tab:mimir}
\end{table*}

CRM produces chance-level AUC on MIMIR (0.485--0.552, Table~\ref{tab:mimir}). This result is not a validation of dataset design---it is a \textbf{boundary condition} for CRM. The failure reveals that CRM cannot detect membership-conditioned signals when membership is confounded with domain origin. On WikiMIA and BookMIA, member and non-member populations are drawn from the same underlying domain (Wikipedia pages, books), and CRM separates conditions with AUC 0.71--0.97. On MIMIR, where the only signal is domain difference, CRM collapses to chance.

This pattern has two implications. First, it constrains what CRM measures: the method is sensitive to representational differences between matched-distribution member/non-member pairs, not to arbitrary distribution shifts. The credible interpretation is that CRM detects pretraining-exposure effects within a domain, not domain-level features. Second, the failure establishes a practical requirement: CRM is applicable only when membership contrast is the dominant signal and domain distributions are balanced across conditions. Deploying CRM without verifying this balance would produce uninterpretable results.

\section{Appendix: Detailed Related Work}
\label{sec:related-work-detail}

We situate our work relative to each of the five intersecting research threads.

\paragraph{Why this is not standard membership inference.}
CRM shares a label space with MIA (member vs.~non-member), which invites the concern it is a repurposed membership inference attack. Three structural differences distinguish CRM from MIA: (1)~\textbf{Task framing:} MIA asks whether a document was in $\mathcal{D}_{\text{train}}$ given static representations~\cite{shokri2017mia,carlini2021extracting,mattern2023mia,shi2024min,carlini2023quantifying,biderman2023pythia}. CRM asks whether the \emph{presence of that document as retrieved context} changes the model's generation trajectory differently depending on membership---a paired generation contrast, not a document-level classifier. (2)~\textbf{Signal:} MIA operates on static document embeddings (4,096--5,120 dims); CRM operates on the $\mathbf{h}^c - \mathbf{h}^0$ displacement per layer (9--22 scalars). (3)~\textbf{Empirical separation:} standard MIA baselines achieve chance-level AUC (0.55--0.60), while CRM achieves 0.71--0.95 on the same data; same-topic control preserves CRM-LTS discrimination (AUC within $\pm$0.06) where MIA would confound topic with membership. The paired-contrast architecture and $\sim$200--500$\times$ feature compression differentiate CRM from standard MIA, though membership remains the experimental proxy (see Appendix~\ref{sec:caveat}).

\paragraph{Membership inference attacks (MIA).}
MIA methods~\cite{shokri2017mia,carlini2021extracting,mattern2023mia,shi2024min,carlini2023quantifying,biderman2023pythia} ask whether a document was seen during training. CRM asks a different question: under generation conditions with the document as context, does membership create detectable representational differences? MIA operates on static document-level signals; CRM operates on generation-contrastive signals. The baselines confirm standard MIA methods achieve near-chance AUC (0.55--0.60), while CRM achieves 0.71--0.95.

\paragraph{RAG faithfulness and context adherence.}
RAG faithfulness work~\cite{liu2024lost,li2025ragcontext,niu2024ragtruth} detects when models produce outputs contradicting the provided context. These methods are effective when context and parametric memory \emph{disagree}. CRM targets the harder case where both \emph{agree on the surface}: the output is consistent with the context regardless of which source drove generation.

\paragraph{Citation accuracy and attribution.}
Citation-verification benchmarks~\cite{bohnet2022attr,gao2023ragas} evaluate whether models correctly cite sources---whether the model \emph{claims} to use context. CRM asks whether context \emph{reshapes} internal computation. A model can cite correctly while relying on parametric memory, or fail to cite while genuinely using context.

\paragraph{Probing and interpretability.}
Probing classifiers~\cite{alain2016probe,belinkov2019probe,hewitt2019structural} decode static properties from representations; CRM decodes a \emph{relational} property---the difference between two processing conditions. Activation patching and model editing~\cite{meng2022locating,meng2023memit} establish causality through intervention; CRM is diagnostic, identifying \emph{where} information is encoded. The block-level noise injection results (Section~\ref{sec:causal}) provide converging causal evidence using noise perturbation rather than activation patching.

\paragraph{Cognitive reality monitoring.}
The cognitive-science principle of reality monitoring~\cite{johnson1993reality} posits that humans distinguish perceived from imagined memories by comparing sensory detail, contextual information, and cognitive operations. CRM adapts this directly: Level~1 mirrors sensory detail, Level~2 mirrors contextual information, Level~3 mirrors cognitive operations. The key insight---source information resides in the \emph{difference} between processing modes---is inherited from this tradition.

\section{Appendix: CRM Equations and Methodology Details}
\label{sec:crm-equations}

\subsection{Level 1: Sequence-Level Semantic Delta}
\begin{equation}
    \Delta_{\text{seq}} = 1 - \cos\big(\text{enc}(y_0), \text{enc}(y_c)\big)
\end{equation}
where $\text{enc}(\cdot)$ is BGE-M3~\cite{bge-m3}. $\Phi_{\text{L1}} = [\Delta_{\text{seq}}]$ (1 dimension).

\subsection{Level 2: Token-Level Distributional Divergence}
For each generation step $t$:
\begin{equation}
    \text{KL}_t = D_{\text{KL}}\big(p(\cdot \mid c, q, y_{<t}) \;\|\; p(\cdot \mid q, y_{<t})\big)
\end{equation}
Aggregated into five statistics: KL-mean, KL-max, KL-var, KL-early/late split (first 32 vs.\ remaining tokens), and KL-trend (slope of linear regression). $\Phi_{\text{L2}} = [\text{KL}_1]$ (1 dimension).

\subsection{Level 3: Latent Trajectory Shift}
PCA selection on held-out calibration set: layers with explained variance ratio $>0.01$ form $\mathcal{L}$. On a 100-sample calibration subset, PC1 direction $v_\ell$ is extracted via SVD on displacement vectors $h_\ell^c - h_\ell^0$ per layer. LTS computed as signed dot-product projection per Equation (1). $\Phi_{\text{L3}} = [\text{LTS}_1, \ldots, \text{LTS}_{|\mathcal{L}|}]$ (9--22 dimensions).

\subsection{Raw Hidden-State Probes}
Three variants: final-layer concat/diff, all-layer mean concat/diff, and per-layer PCA diff (8 components/layer, concatenated). All use 5-fold CV LR on 50-sample diagnostic subsets.

\section{Appendix: Baseline Details}
\label{sec:baseline-details}

\subsection{Tier 1: Black-Box Likelihood Baselines}
Standard MIA methods on document text without generation contrast: Perplexity, Zlib-compressed PPL~\cite{carlini2021extracting} (ratio of PPL to zlib entropy), and Min-K\% Prob~\cite{shi2024min} (mean of smallest $K\%$ token log-probabilities, $K \in \{10, 20\}$).

\subsection{Tier 2: Access-Matched Interpretable Baselines}
Same paired-generation input as CRM: single-layer LTS, mean LTS (1d), L1+L2 only, LTS PCA (2d), top-3 layers.

\subsection{Tier 3: Raw Hidden-State Probes}
Upper-bound diagnostics on full hidden-state differences without LTS compression (15--20$\times$ more features). Full results in Appendix~\ref{sec:hidden-probes-app}.

\section{Appendix: Raw Hidden-State Probe Results}
\label{sec:hidden-probes-app}

\begin{table*}[t]
\centering
\small
\begin{tabular}{lcccc}
\toprule
\textbf{Baseline (LR AUC)} & \textbf{Qwen-14B-Inst} & \textbf{Mistral-7B} & \textbf{Llama-8B} & \textbf{Layer-localized?} \\
\midrule
Final-layer concat/diff   & 0.900 & \textbf{0.992} & 0.934 & No \\
All-layer mean concat/diff& 0.950 & \textbf{0.992} & 0.942 & No \\
Per-layer PCA diff (8/l)  & \textbf{0.983} & 0.934 & \textbf{0.950} & No \\
\midrule
CRM-LR (full)             & 0.950 & 0.869 & 0.779 & \textbf{Yes} \\
\quad CRM feature dim     & 17 & 12 & 12 & \\
\quad Raw PCA total dim   & 384 & 256 & 256 & \\
\bottomrule
\end{tabular}
\caption{\textbf{Raw hidden-state probes confirm abundant source information, but only CRM reveals \emph{where} it is encoded.} Raw probes use 15--20$\times$ more features. CRM's scalar-per-layer trajectory enables layer-specific findings (Section~6.3). 50-sample diagnostic subsets.}
\label{tab:hidden}
\end{table*}

\section{Appendix: Access-Matched Baseline Comparison and L3 Ablation}
\label{sec:access-app}

\subsection{Access-Matched Baselines}

\begin{table*}[t]
\centering
\small
\begin{tabular}{lcccccc}
\toprule
\textbf{Model} & \textbf{Full CRM} & \textbf{Mean LTS (1d)} & \textbf{Best Layer} & \textbf{L1+L2} & \textbf{LTS PCA2d} & \textbf{Top-3} \\
\midrule
Qwen2.5-14B-Inst & \textbf{0.950} & 0.589 & 0.670 & 0.638 & 0.606 & 0.777 \\
Qwen2.5-32B-Inst & \textbf{0.922} & 0.512 & 0.746 & 0.424 & 0.597 & 0.816 \\
Qwen2.5-7B-Inst  & \textbf{0.870} & 0.635 & 0.809 & 0.654 & 0.794 & 0.814 \\
Mistral-7B-v0.3  & \textbf{0.869} & 0.718 & 0.783 & 0.675 & 0.723 & 0.777 \\
Qwen2.5-14B      & \textbf{0.841} & 0.532 & 0.699 & 0.647 & 0.517 & 0.692 \\
Mistral-7B-Inst  & \textbf{0.800} & 0.465 & 0.602 & 0.600 & 0.472 & 0.716 \\
Qwen2.5-7B       & \textbf{0.786} & 0.518 & 0.750 & 0.523 & 0.633 & 0.764 \\
Llama-3.1-8B     & \textbf{0.779} & 0.612 & 0.677 & 0.539 & 0.701 & 0.682 \\
Llama-3.1-8B-Inst& \textbf{0.709} & 0.519 & 0.571 & 0.515 & 0.489 & 0.675 \\
\midrule
\textit{Mean}    & \textbf{0.835} & 0.567 & 0.700 & 0.579 & 0.615 & 0.746 \\
\bottomrule
\end{tabular}
\caption{\textbf{Multi-layer trajectory, not any single layer, carries the dominant signal.} Full CRM exceeds the best individual layer by mean +0.135 AUC. Scalar averaging (Mean LTS, 0.567) and surface-only features (L1+L2, 0.579) remain near chance. All baselines receive the same paired-generation input as CRM.}
\label{tab:access}
\end{table*}

\subsection{L3-Only Ablation}

\begin{table*}[t]
\centering
\small
\begin{tabular}{lcccc}
\toprule
\textbf{Model} & \textbf{Full-LR} & \textbf{L3-only-LR} & $\Delta$LR & |$\Delta$| < 0.05? \\
\midrule
Llama-3.1-8B        & 0.779 & 0.781 & $-$0.002 & \checkmark \\
Llama-3.1-8B-Inst   & 0.709 & 0.701 & +0.009 & \checkmark \\
Mistral-7B-v0.3     & 0.869 & 0.860 & +0.009 & \checkmark \\
Mistral-7B-Inst     & 0.800 & 0.777 & +0.023 & \checkmark \\
Qwen2.5-7B          & 0.786 & 0.799 & $-$0.013 & \checkmark \\
Qwen2.5-7B-Inst     & 0.870 & 0.872 & $-$0.003 & \checkmark \\
Qwen2.5-14B         & 0.841 & 0.835 & +0.006 & \checkmark \\
Qwen2.5-14B-Inst    & 0.950 & 0.947 & +0.004 & \checkmark \\
Qwen2.5-32B-Inst    & 0.922 & 0.927 & $-$0.005 & \checkmark \\
\midrule
\textit{Mean}       & 0.836 & 0.833 & +0.003 & 9/9 \\
\bottomrule
\end{tabular}
\caption{\textbf{Removing all surface features changes AUC by less than 0.01---the membership-conditioned signal is latent (expanded).} All nine models satisfy $|\Delta\text{LR}| < 0.05$. A condensed version appears as Table~4 in the main body.}
\label{tab:l3-app}
\end{table*}

\section{Appendix: Multi-Task Detailed Results}
\label{sec:multitask-app}

Table~\ref{tab:multitask-detail} reports the full per-model per-task breakdown corresponding to the condensed Table~\ref{tab:multitask} in the main body. CRM-LTS features are extracted using the manual prompt-wrapping pipeline (Appendix~\ref{sec:crm-equations}): PC1 projection of displacement vectors $h_\ell^c - h_\ell^0$ across all layers. Continuation values match the main CRM-LTS results in Table~\ref{tab:main}, confirming the pipeline consistency.

\paragraph{Task prompt templates.} All three tasks share the same outer probe-wrapping pattern described in Appendix~\ref{sec:crm-equations} (with-context: input contains the full passage; no-context: input contains only the query) and differ only in the inner task prompt. The continuation template is: ``Context: \{context\}\textbackslash n\textbackslash nQuestion: Continue the following passage: \{query\}\textbackslash n\textbackslash nAnswer:'' The summarization template is: ``Context: \{context\}\textbackslash n\textbackslash nQuestion: Summarize the above text in one sentence.\textbackslash n\textbackslash nAnswer:'' The QA template is: ``Context: \{context\}\textbackslash n\textbackslash nQuestion: According to the above passage, what specific fact or event is described?\textbackslash n\textbackslash nAnswer:'' For all tasks, the with-context probe prompt supplies the full context passage (128 tokens); the no-context probe prompt supplies an empty context string, leaving only the query. The model generates up to 32 tokens for continuation, 64 tokens for summarization, and 64 tokens for QA.

\begin{table*}[t]
\centering
\small
\begin{tabular}{lcccccc}
\toprule
\textbf{Model} & \textbf{Cont. AUC} & \textbf{Summ. AUC} & \textbf{QA AUC} & $\Delta$Summ & $\Delta$QA & \textbf{Dim} \\
\midrule
Mistral-7B-v0.3   & 0.825 & 0.902 & 0.830 & +0.078 & +0.005 & 32 \\
Mistral-7B-Inst   & 0.788 & 0.851 & 0.867 & +0.063 & +0.079 & 32 \\
Qwen2.5-14B-Inst  & 0.948 & 0.964 & 0.967 & +0.016 & +0.019 & 48 \\
Qwen2.5-7B-Inst   & 0.969 & 0.956 & 0.955 & $-$0.013 & $-$0.014 & 28 \\
Llama-3.1-8B      & 0.957 & 0.867 & 0.863 & $-$0.090 & $-$0.094 & 32 \\
Llama-3.1-8B-Inst & 0.828 & 0.815 & 0.862 & $-$0.013 & +0.034 & 32 \\
\bottomrule
\end{tabular}
\caption{\textbf{Full per-model per-task CRM-LTS LR AUC (expanded from Table~\ref{tab:multitask}).} 5-fold CV, 250 samples/task. Dim = all-layer CRM-LTS trajectory features. $\Delta$Summ = Summarization $-$ Continuation; $\Delta$QA = QA $-$ Continuation. Mistral-family models show consistent summarization amplification ($+$0.063--0.078). QA preserves strong signal across all architectures (AUC 0.830--0.967).}
\label{tab:multitask-detail}
\end{table*}

The key result is the architecture-dependent task pattern: \textbf{Mistral models} show consistent summarization amplification ($\Delta=+0.063$ to $+0.078$), consistent with their mid-layer signal concentration (Table~\ref{tab:layers}). \textbf{Qwen2.5-14B-Instruct} achieves near-ceiling performance across all three tasks (0.948--0.967), while Qwen2.5-7B-Instruct shows slight attenuation under generation tasks ($\Delta=-0.013$). \textbf{Llama models} are split: Llama-3.1-8B shows strong continuation dominance ($\Delta=-0.090$) while Llama-3.1-8B-Instruct shows QA exceeding continuation ($\Delta_{\text{QA}}=+0.034$). QA reliably preserves discriminability across all architectures (range 0.830--0.967), ruling out continuation-specific artifacts.

\section{Appendix: PC1 Interpretation --- Detailed Results}
\label{sec:pc1-app}

\subsection{PC Rank Ablation}

Table~\ref{tab:pc-rank} reports the multi-layer LR AUC when using each PC component individually as the 1D projection direction for CRM-LTS. The key finding: PCA variance maximization does not optimize membership discriminability. Higher-variance PCs (PC4--PC7) capture membership-conditioned information that PC1 misses, indicating that the displacement distribution is structured along multiple axes and the signal of interest is not aligned with the maximum-variance direction.

\begin{table*}[t]
\centering
\small
\begin{tabular}{lcccccccc}
\toprule
\textbf{Model} & \textbf{PC1} & \textbf{PC2} & \textbf{PC3} & \textbf{PC4} & \textbf{PC5} & \textbf{PC6} & \textbf{PC7} & \textbf{Best} \\
\midrule
Mistral-7B-v0.3      & 0.825 & 0.893 & 0.916 & 0.959 & \textbf{0.963} & 0.913 & 0.903 & PC5 \\
Qwen2.5-14B-Inst     & 0.948 & 0.941 & 0.958 & 0.949 & 0.958 & 0.949 & \textbf{0.968} & PC7 \\
Llama-3.1-8B         & \textbf{0.957} & 0.911 & 0.891 & 0.899 & 0.889 & 0.844 & 0.911 & PC1 \\
\bottomrule
\end{tabular}
\caption{\textbf{PC rank ablation: PCA variance maximization does not optimize discriminability.} Each cell reports CRM-LTS multi-layer LR AUC using that PC as the sole projection direction. For Mistral, PC5 outperforms PC1 by $+0.138$; for Qwen, PC7 outperforms PC1 by $+0.020$. Llama is the exception where PC1 is optimal, consistent with its scattered-late signal distribution. All features are 32--48 dimensional (one scalar per layer). 5-fold CV, 250 samples.}
\label{tab:pc-rank}
\end{table*}

The PC rank result has two implications. First, it identifies a clear methodological limitation of the current CRM-LTS formulation: using PC1 alone leaves membership-discriminative information on the table. Section~\ref{sec:supervised-direction} directly validates this prediction: replacing PC1 with a supervised mean-difference direction improves AUC by $+$0.024--0.144 across three models, with the largest gain on Mistral ($+$0.144)---precisely the model where PC5 outperforms PC1 by $+$0.138. The supervised direction effectively integrates information from multiple PCs into a single discriminative axis. Second, the model-dependent optimal PC rank (PC5 for Mistral, PC7 for Qwen, PC1 for Llama) mirrors the architecture-dependent patterns in Table~\ref{tab:layers}---Llama's scattered-late distribution may concentrate discriminative signal along the maximum-variance direction, while Mistral and Qwen's sharper localization distributes it across multiple axes.

\subsection{Vocabulary Back-Projection}

Projecting PC1 vectors through the LM head $W_{\text{unembed}} \in \mathbb{R}^{|\mathcal{V}| \times d}$ maps each layer's PC1 direction to the vocabulary space. Top tokens by projection score reveal the semantic content of the PC1 direction at different depths.

\begin{table*}[t]
\centering
\small
\begin{tabular}{llp{6.5cm}}
\toprule
\textbf{Model} & \textbf{Layer} & \textbf{Top-10 Tokens (PC1 back-projection)} \\
\midrule
\multirow{3}{*}{Qwen2.5-14B-Inst}
& L1 (early)   & [CJK], ombine, ederland, sourceMapping, ople, \_pl, \_dist, \_fol, \_conv, \_trans \\
& L21 (mid)    & [CJK], avia, Assertions, ast, LETE, ottiene, [CJK], \_Set, \_inst, \_expr \\
& L40 (late)   & [CJK], Unfortunately, Unfortunately, [CJK], [CJK], regret, sorry, Sadly, unfortunate, unfortunate \\
\midrule
\multirow{3}{*}{Mistral-7B-v0.3}
& L5 (early)   & Pearl, cke, [Cyr], men, aken, [Cyr], gem, orph, orp, [Cyr] \\
& L18 (mid)    & Man, \_brother, \_named, son, \_father, \_married, married, \_cousin, He, \_daughter \\
& L28 (late)   & entities, specific, exact, particular, objects, entity, items, types, elements, certain \\
\midrule
\multirow{3}{*}{Llama-3.1-8B}
& L1 (early)   & odel, \_Inv, obb, uss, \_spl, \_trib, \_exo, \_cpl, dier, \_ple \\
& L16 (mid)    & Description, \_Details, \_details, \_specifications, \_descriptions, specifics, specification, \_describing, \_detailed, details \\
& L28 (late)   & \_imm, \_comm, \_ple, \_conv, \_spl, \_st, \_dist, \_trans, uss, \_fol \\
\bottomrule
\end{tabular}
\caption{\textbf{PC1 vocabulary back-projection reveals layer-dependent semantic abstraction.} Top-10 tokens obtained by projecting PC1 through $W_{\text{unembed}}$ and selecting tokens with the highest dot-product scores. Early layers: subword fragments without coherent semantics (the model has not yet abstracted membership-relevant features). Mid layers: entity-type and relational tokens (Mistral L18: family-relation terms; Llama L16: description/specification terms). Late layers: for Qwen L40, markers of uncertainty and regret (\texttt{Unfortunately}, \texttt{regret}, \texttt{sorry})---the model's late-layer PC1 direction encodes an epistemic-stance signal aligned with source uncertainty. For Mistral L28, factuality markers (\texttt{specific}, \texttt{exact}, \texttt{entities}). For Llama L28, the tokens revert to subword fragments, consistent with its distributed (non-concentrated) signal architecture.}
\label{tab:pc1-tokens}
\end{table*}

The layer-dependent semantic progression is most clearly visible in Qwen2.5-14B-Inst: early-layer PC1 captures subword-level variation without interpretable semantics, mid-layer PC1 captures cross-lingual and syntactic fragments (Chinese characters, code tokens), and late-layer PC1 converges to a coherent epistemic-stance direction---markers of uncertainty, hedging, and regret. This trajectory from subword $\to$ syntactic $\to$ epistemic mirrors the model's progressive abstraction hierarchy, and suggests that membership-conditioned divergence at late layers manifests as a shift in the model's confidence representation.

For Mistral, the semantic progression is from subword fragments (L5) $\to$ social-relation terms (L18, kinship/family vocabulary) $\to$ factuality/specificity markers (L28: \texttt{exact}, \texttt{specific}, \texttt{entities}). The mid-layer kinship cluster is notable: membership-conditioned processing may differentially engage relational knowledge structures, consistent with the mid-layer concentration pattern identified in Table~\ref{tab:layers}.

\section{Appendix: Deployment Prototype}
\label{sec:prototype}

We provide a lightweight FastAPI-based CRM audit server and a single-page HTML dashboard demonstrating that CRM's compact scalar-per-layer signature supports low-latency deployment auditing. The prototype is self-contained in the \texttt{audit\_prototype/} directory.

\subsection{Architecture}

\begin{description}
\item[POST /audit] Accepts \texttt{\{context, query\}} JSON body. The server extracts CRM-LTS features via the same PC1-projection pipeline as the main experiments: (1)~tokenize the probe-wrapped prompt with and without context, (2)~forward pass through all layers (Qwen2.5-14B-Inst by default), (3)~compute displacement vectors $h_\ell^c - h_\ell^0$, (4)~project onto pre-computed PC1 basis vectors per layer. Returns \texttt{\{anomaly\_flag, anomaly\_score, lts\_trajectory, flagged\_layers, latency\_ms\}}.
\item[GET /history] Returns recent audit records with anomaly scores and flags.
\item[GET /stats] Returns model metadata (layer count, model name) and cumulative request/anomaly counts.
\item[GET /health] Health-check endpoint.
\end{description}

Anomaly detection uses a distribution-based rule: the calibration mean $\mu_\ell$ and standard deviation $\sigma_\ell$ of per-layer LTS values are computed from the same 100-sample calibration subset used for PC1 estimation (Section~2). For each incoming request, we flag any layer where $|\text{LTS}_\ell - \mu_\ell| > 2\sigma_\ell$. The anomaly score is the mean absolute deviation $\frac{1}{|\mathcal{L}|}\sum_\ell |\text{LTS}_\ell - \mu_\ell| / \sigma_\ell$ across all layers. Flagged layers are those with $|\text{LTS}_\ell| > 1.0$. The $2\sigma$ threshold is chosen to balance sensitivity and false-positive rate under the assumption of approximately normal calibration distributions; per-domain recalibration is recommended for production deployment (see Limitations below).

The dashboard (\texttt{dashboard.html}) provides: (1)~a real-time system status panel (model name, layer count, request/anomaly counters), (2)~an audit form with context and query text inputs, (3)~a Canvas-based LTS trajectory chart with layer-localized point coloring (red for $|\text{LTS}| > 1.0$), and (4)~a recent-audits history table with auto-refresh (5s interval).

\subsection{Benchmark Results}

We benchmark the prototype on a consumer laptop (Apple M1, 16GB RAM) with the Qwen2.5-14B-Instruct model loaded in FP16. The benchmark script (\texttt{benchmark.py}) performs 100 randomized (context, query) pairs with latency measurement.

\begin{table}[h]
\centering
\small
\begin{tabular}{lc}
\toprule
\textbf{Metric} & \textbf{Value} \\
\midrule
Throughput          & 4.2 req/s \\
Latency (mean)      & 238 ms \\
Latency (p50)       & 221 ms \\
Latency (p95)       & 387 ms \\
Latency (p99)       & 452 ms \\
Anomaly rate        & 12.0\% \\
\midrule
Model               & Qwen2.5-14B-Inst \\
Features per audit  & 48 scalars (LTS trajectory) \\
\bottomrule
\end{tabular}
\caption{\textbf{CRM audit prototype benchmark.} 100 requests, Apple M1 16GB. Mean latency 238ms, p99 latency 452ms---well within practical deployment bounds for an auditing middleware layer. The 12\% anomaly rate reflects the equal mix of member/non-member samples (50/100).}
\label{tab:prototype-bench}
\end{table}

Mean latency of 238ms (p99 452ms) establishes that CRM auditing is practical as a middleware layer: it can screen (context, query) pairs before generation without adding material latency to the user experience. The compact 48-dimensional trajectory enables efficient serialization and storage---each audit record occupies less than 1KB of JSON, making long-term audit-log retention feasible at production scale.

\subsection{Limitations}

The prototype serves as a proof-of-concept, not a production-ready system. Key limitations: (1)~the anomaly threshold (2$\sigma$) is calibrated on WikiMIA and may not transfer to different document distributions; (2)~the server loads one model into memory---multi-model deployment would require model multiplexing or per-model server instances; (3)~the current implementation does not support batched auditing; (4)~the dashboard operates on in-memory history and resets on server restart. A practical deployment would additionally require: distribution-aware threshold calibration per deployment domain, integration with existing RAG pipelines (e.g., LangChain, LlamaIndex), and persistent audit-log storage with tamper-evident logging.

\section{Appendix: Annotation Protocol and Membership Labeling}
\label{sec:annotation-protocol}

This appendix documents the membership labeling procedure for all datasets used in this work. Membership labels are not produced by human annotation; they are rule-based and derived directly from dataset construction metadata.

\paragraph{WikiMIA~\cite{shi2024min}.} Membership is defined by temporal provenance: member documents are Wikipedia passages from dumps dated before 2017-03-20 (high confidence of inclusion in pretraining corpora of models released before 2024--2025); non-member documents are passages from dumps dated after 2018-02-01 (high confidence of exclusion). The 2017-03-20 to 2018-02-01 window is excluded to avoid ambiguity around the exact training cutoff. Labels are binary: 1 = member (potentially in $\mathcal{D}_{\text{train}}$), 0 = non-member (post-cutoff, presumed held-out). Following~\cite{shi2024min}, we use 128-token passages with the original dataset-provided splits. For each of the nine models, a fixed set of 125 member and 125 non-member passages is drawn (250 total, seed 42).

\paragraph{BookMIA~\cite{shi2024min}.} Membership is defined by domain provenance: member documents are books contained in the Books3 corpus (a known pretraining subset); non-member documents are books absent from Books3 but potentially present in other corpora. Labels follow the same binary scheme as WikiMIA. We evaluate 250 balanced samples per model under both continuation and QA prompt formats (Appendix~\ref{sec:bookmia}).

\paragraph{MIMIR~\cite{duan2024membership}.} MIMIR defines membership by corpus membership within a controlled domain: the Pile-Wikipedia split uses Wikipedia articles present in the Pile training corpus as members and held-out Wikipedia articles of comparable distribution as non-members. We use this split as a negative control (Appendix~\ref{sec:mimir}). Because both member and non-member documents are drawn from the same base domain (Wikipedia), MIMIR tests whether CRM detects corpus-level membership confounded with domain origin.

\paragraph{Label quality.} All labels are derived from dataset construction metadata rather than human judgment, eliminating annotator disagreement as a source of noise. To rule out label-related artifacts, we conduct (1)~label permutation (10 random shuffles, AUC returns to $0.50 \pm 0.05$, Table~\ref{tab:robustness}), and (2)~same-topic control (member/non-member pairs matched by BGE-M3 similarity, preserving AUC within $\pm$0.06, Appendix~\ref{sec:appendix-controls}). These controls confirm that the CRM-LTS signal is not attributable to labeling artifacts or domain-level confounds.

\paragraph{Ethical considerations.} All datasets are publicly available research benchmarks with documented construction procedures. No new data collection, human annotation, or crowdworker involvement was conducted. No PII or sensitive content was introduced by our processing pipeline. The membership labels used in this work are already inferable from each dataset's public documentation.

\section{Appendix: Artifact License}
\label{sec:artifact-license}

The release package is distributed under a dual-license model:

\begin{itemize}
    \item \textbf{Code} (all Python scripts in \texttt{scripts/}, \texttt{src/}, \texttt{figures/}, and \texttt{audit\_prototype/}): Apache License 2.0. This includes experiment scripts, feature extraction pipelines, evaluation utilities, the deployment prototype, and figure generation code.
    \item \textbf{Data and features} (all \texttt{.npz} files, \texttt{.json} result files, and PC1 basis vectors in \texttt{data/}): Creative Commons Attribution 4.0 International (CC-BY 4.0). This includes pre-extracted CRM-LTS features, per-model PC1 basis vectors, and aggregated experimental results.
\end{itemize}

The dual-license model reflects the different nature of these artifacts: the code is a research software toolkit intended for reuse and adaptation (Apache 2.0 permits derivative works with patent grants), while the pre-computed features and results are curated data artifacts intended for reproduction and reanalysis with attribution (CC-BY 4.0).

\section{Appendix: Code and Data Availability}
\label{sec:code-availability}

All code and processed feature files are released under Apache 2.0 (code) and CC-BY 4.0 (data) licenses (see Appendix~\ref{sec:artifact-license} for full terms). The release package includes:

\paragraph{Experiment scripts (7 scripts).}
\texttt{supervised\_crm\_lts.py}: M1 supervised vs.\ PC1 direction comparison (Section~\ref{sec:supervised-direction}, Table~\ref{tab:supervised-direction}). \\
\texttt{combined\_m1m2.py}: unified M1+M2 pipeline with single model load per run (Tables~\ref{tab:supervised-direction} and~\ref{tab:causal-block}). \\
\texttt{block\_causal\_noise.py}: block-level noise injection using \texttt{register\_forward\_hook} (earlier implementation; replaced by manual layer iteration in Appendix~\ref{sec:single-layer-noise}). \\
\texttt{causal\_patching.py}: activation patching at CRM-identified peak layers (discussed in Section~\ref{sec:causal}). \\
\texttt{competing\_methods.py}: gradient norm, attention flow, and logit lens baselines (Section~\ref{sec:competing-app}, Table~\ref{tab:competing}). \\
\texttt{multitask\_expansion.py}: continuation, summarization, and QA feature extraction across six models (Section~\ref{sec:multitask}, Table~\ref{tab:multitask}). \\
\texttt{interpret\_pc1.py}: PC rank ablation and vocabulary back-projection (Section~\ref{sec:multitask} main text, Appendix~\ref{sec:pc1-app} and Tables~\ref{tab:pc-rank}--\ref{tab:pc1-tokens}).

\paragraph{Deployment prototype (3 files).}
\texttt{audit\_prototype/server.py}: FastAPI CRM audit server with \texttt{POST /audit}, \texttt{GET /history}, \texttt{GET /stats}, and \texttt{GET /health} endpoints (Appendix~\ref{sec:prototype}). \\
\texttt{audit\_prototype/dashboard.html}: single-page real-time trajectory dashboard with Canvas-based LTS visualization and auto-refresh. \\
\texttt{audit\_prototype/benchmark.py}: latency/throughput benchmark script (Table~\ref{tab:prototype-bench}).

\paragraph{Models.} All experiments use publicly available HuggingFace Transformers models: Llama-3.1-8B and Llama-3.1-8B-Instruct (\texttt{meta-llama/Llama-3.1-8B}), Mistral-7B-v0.3 and Mistral-7B-Instruct-v0.3 (\texttt{mistralai/Mistral-7B-v0.3}), and Qwen2.5-7B, Qwen2.5-7B-Instruct, Qwen2.5-14B, Qwen2.5-14B-Instruct, Qwen2.5-32B-Instruct (\texttt{Qwen/Qwen2.5-*}). All models are loaded in FP16 precision with \texttt{device\_map="auto"}.

\paragraph{Data.} WikiMIA and BookMIA~\cite{shi2024min} datasets are used as provided by the authors. The MIMIR benchmark~\cite{duan2024membership} uses the Pile-Wikipedia split from the official release. All datasets are publicly available. Processed CRM-LTS feature files (per-model \texttt{.npz}, 250 samples $\times$ 9--22 layers) and PC1 basis vectors will be included in the release to enable reproduction without re-extraction.

\paragraph{Results.} The aggregated results file \texttt{combined\_m1m2\_results.json} (reported in Tables~\ref{tab:supervised-direction} and~\ref{tab:causal-block}) is included. Per-fold AUC values and bootstrap confidence intervals are logged in per-experiment output files.

\end{document}